\documentclass[lettersize,journal]{IEEEtran}
\usepackage{amsmath,amsfonts}
\usepackage{amssymb}
\usepackage{array}
\usepackage[caption=false,font=normalsize,labelfont=sf,textfont=sf]{subfig}
\usepackage{textcomp}
\usepackage{stfloats}
\usepackage{url}
\usepackage{verbatim}
\usepackage{graphicx}
\usepackage{cite}

\usepackage[linesnumbered,ruled,vlined]{algorithm2e}
\usepackage{algpseudocode}

\usepackage{ntheorem}
\newtheorem{myAssume}{Assumption}
\newtheorem{myDef}{Definition}
\newtheorem{myTheo}{Theorem}
\newtheorem*{myProof}{Proof}
\newtheorem{myLemma}{Lemma}

\newtheorem{myCorollary}{Corollary}

\usepackage{amsmath, bm}
\usepackage{amsfonts}
\usepackage{enumitem}
\usepackage{float} 
\usepackage{multirow,multicol}
\usepackage{booktabs}
\usepackage{xcolor}
\usepackage{hyperref}

\hypersetup{%
  colorlinks=true,%
  linkcolor=red,
  citecolor=blue,
  urlcolor=blue,
  bookmarksnumbered=true,%
  bookmarksopen=true}

\usepackage{listings}
\usepackage{color}

\definecolor{dkgreen}{rgb}{0,0.6,0}
\definecolor{gray}{rgb}{0.5,0.5,0.5}
\definecolor{mauve}{rgb}{0.58,0,0.82}

\begin{document}

\title{Learning to Substitute Components for Compositional Generalization}

\author{Zhaoyi Li, Gangwei Jiang, Chenwang Wu, Ying Wei, Defu Lian, and Enhong Chen,~\IEEEmembership{Fellow,~IEEE}

\thanks{Corresponding authors: Ying Wei and Defu Lian.}
\thanks{Zhaoyi Li, Gangwei Jiang, Defu Lian and Enhong Chen are affiliated with the School of Computer Science and Technology, University of Science and Technology of China, Hefei, Anhui 230026, P.R. China (e-mail: lizhaoyi777@mail.ustc.edu.cn, gwjiang@mail.ustc.edu.cn, liandefu@ustc.edu.cn, cheneh@ustc.edu.cn).}
\thanks{Chenwang Wu is affiliated with the Department of Computer Science, Hong Kong Baptist University, Kowloon, Kowloon Tong 999077, Hong Kong (e-mail:wcw1996@mail.ustc.edu.cn).}
\thanks{Ying Wei is affiliated with the College of Computer Science and Technology, Zhejiang University, Hangzhou, Zhejiang 310058, P.R. China (e-mail:ying.wei@zju.edu.cn).}}

\markboth{Journal of \LaTeX\ Class Files,~Vol.~14, No.~8, August~2021}%
{Shell \MakeLowercase{\textit{et al.}}: Compositional Generalization via Learning Component Substitution}


\maketitle

\begin{abstract}
Despite the rising prevalence of neural language models, recent empirical evidence suggests their deficiency in compositional generalization. One of the current de-facto solutions to this problem is compositional data augmentation, which aims to introduce additional compositional inductive bias. However, existing handcrafted augmentation strategies offer limited improvement when systematic generalization of neural language models requires multi-grained compositional bias (i.e., not limited to either lexical or structural biases alone) or when training sentences have an imbalanced difficulty distribution.
To address these challenges, we first propose a novel compositional augmentation strategy called \textbf{Comp}onent \textbf{Sub}stitution (CompSub), which enables multi-grained composition of substantial substructures across the entire training set. Furthermore, we introduce the \textbf{L}earning \textbf{C}omponent \textbf{S}ubstitution (LCS) framework. This framework empowers the learning of component substitution probabilities in CompSub in an end-to-end manner by maximizing the loss of neural language models, thereby prioritizing challenging compositions with elusive concepts and novel contexts.
We extend the key ideas of CompSub and LCS to the recently emerging in-context learning scenarios of pre-trained large language models (LLMs), proposing the LCS-ICL algorithm to enhance the few-shot compositional generalization of state-of-the-art (SOTA) LLMs. Theoretically, we provide insights into why applying our algorithms to language models can improve compositional generalization performance. Empirically, our results on four standard compositional generalization benchmarks—SCAN, COGS, GeoQuery, and COGS-QL—demonstrate the superiority of CompSub, LCS, and LCS-ICL, with improvements of up to 66.5\%, 10.3\%, 1.4\%, and 8.8\%, respectively.
\end{abstract}
\section{Introduction}
\label{sec:intro}
\begin{figure}[h]
\centering 
\includegraphics[width=0.47\textwidth]{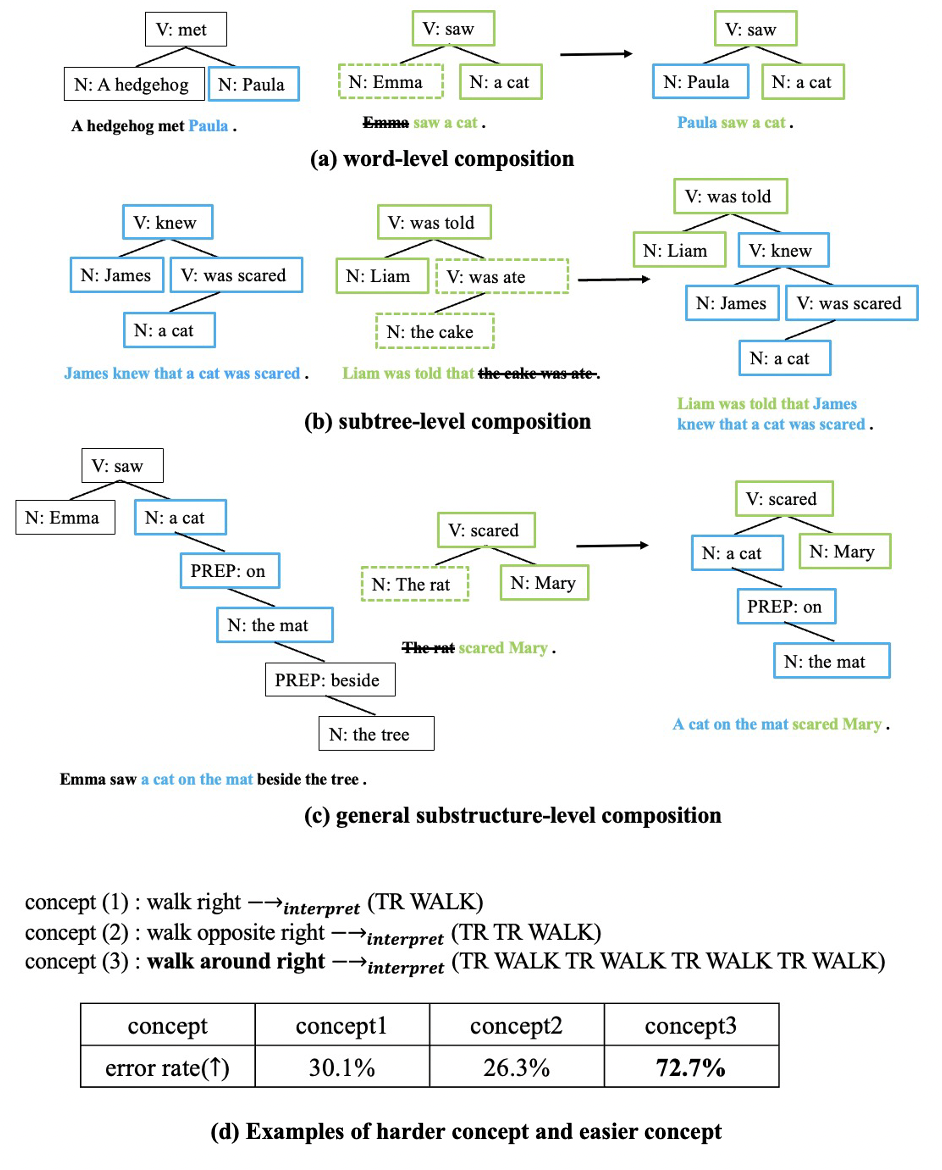}
\caption{(a), (b) and (c) illustrate three distinct compositional generalization types in COGS~\cite{cogs}, which require word-level, subtree-level and general substructure-level recombinations of training data, respectively. Besides, (d) shows concepts in distinct difficulty in the SCAN~\cite{scan} dataset, where the interpretation of \emph{walk around right} is much more complex than that of the other two concepts.}
\label{figure:intro}
\vspace{-0.1in}
\end{figure}

The secret for human beings to learning so quickly with little supervision has been demonstrated to be associated with the powerful ability of \emph{compositional generalization}~\cite{doi:10.1073/pnas.2205582119,Lake2023,Riveland2024,yagcioglu-etal-2024-sequential}, being capable of producing an infinite number of novel combinations on the basis of known components~\cite{chomsky,Lake2023}.

In stark contrast, a large body of recent evidence suggests that current state-of-the-art neural language models (including LSTM-based Sequence-to-Sequence models (Seq-to-Seq)~\cite{LSTM, seq2seq,attention}, Transformers~\cite{transformer}, pre-trained language models (PLMs)~\cite{bart,t5} and large language models (LLMs)~\cite{gpt3,touvron2023llama2openfoundation,dubey2024llama3herdmodels}) lack of adequate power for compositional generalization (\emph{a.k.a.,} systematic generalization)~\cite{scan, cogs, MCD, pretrain.vs.special,an-etal-2023-context}. 
For instance, a model which has observed the two training sentences of ``\emph{look opposite right} twice and jump right thrice'' and ``\emph{walk around right} and run twice'' likely fails to understand the testing sentence of ``\emph{walk around right} twice and jump right thrice''.
Sharpening the compositional generalization ability of neural language models is beyond important to fill the gap with human-like natural language understanding, catalyzing not only better performances but also fewer expensive annotations.

Inspired by the tight relationship between compositionality and group-equivariance of neural models~\cite{permutation-eq,lexsym,Basu2022EquiTuningGE}, a series of compositional data augmentation solutions have made great strides via injecting compositional inductive bias into neural language models~\cite{geca, seqmix, lexsym, subs, mutual}. 
The key idea behind compositional data augmentation is to substitute a part in one original training example with a part from another training example, thus composing a novel example that complements the training data with compositional bias. 

Introducing comprehensive enough compositional bias to embrace a diversity of testing tasks, however, is not trivial. 
First, the ``part''\footnote{In this paper, We use the words of ``part'', ``concept'', ``span'' and ``component'' later interchangeably.} to be substituted out and in is expected to be in multiple levels, ranging from words~\cite{lexsym} in Figure~\ref{figure:intro}(a), to complete substrees~\cite{subs} in Figure~\ref{figure:intro}(b), to more general substructures in Figure~\ref{figure:intro}(c). How to develop an augmentation method that flexibly accommodates multiple levels of parts remains an open question.
Second, the ``parts'' are uneven in their difficulty levels. 
As shown in Figure~\ref{figure:intro}(d), though the numbers of both training and testing sentences containing the three concepts in the SCAN MCD split~\cite{MCD} are comparable and we have applied compositional data augmentation via the proposed CompSub (which will be detailed later in Section~\ref{sec:spansub}), the predicted error rates of testing sentences grouped by the three concepts still differ quite significantly, which is in alignment with the observations in ~\cite{bogin-etal-2022-unobserved}. There is an urgent need to augment with difficulty awareness and allow more compositions on the challenging concepts (e.g., concept 3 in Figure~\ref{figure:intro}(d), ``walk around right'', is supposed to be mapped to a more complex structure in comparison with the other two concepts ``walk right'' and ``walk opposite right'').

In the previous conference version of our paper~\cite{li-etal-2023-learning}, we propose two methods to address the two aforementioned challenges. 
We first introduce a novel compositional data augmentation scheme, dubbed \textbf{Comp}onent \textbf{Sub}stitution (CompSub). This method substitutes a \emph{component} in a training sentence with one in another sentence. Here a \emph{component} refers to a consecutive fragment of tokens that encompasses all multi-grained possibilities of a word, a subtree, as well as a more general substructure (e.g., ``a cat on the mat'' in Figure~\ref{figure:intro}(c)). The core of CompSub lies in the extraction of such components and the identification of exchangeable components. We define the exchangeability of components based on the syntactic equivalence of their first and last tokens.
However, purely random component substitution in CompSub overlooks the imbalanced difficulty distribution of compositions and hence may lead to suboptimal results. To address this challenge, we propose the \textbf{L}earning \textbf{C}omponent \textbf{S}ubstitution (LCS) framework. This framework includes a LCS augmenter, which is a differentiable version of CompSub with all substitution actions equipped with probabilities. By training down-stream neural language models to evaluate the difficulty of various components (i.e., the concepts in Figure~\ref{figure:intro}(d)) and maximizing their losses, the LCS framework seeks to train the LCS augmenter to prioritize substitution actions that introduce challenging compositions involving elusive components and novel contexts.

In this paper, we additionally provide significant extensions based on the conference paper~\cite{li-etal-2023-learning}. Specifically: 
(1) We offer theoretical insights and analysis for our proposed CompSub and LCS algorithms (Section~\ref{sec:th}). We prove that using our compositional data augmentation method, CompSub, is equivalent to imposing an additional regularization term on the original optimization objective. This highly encourages language models to achieve invariant learning~\cite{arjovsky2020invariantriskminimization,lyle2020benefitsinvarianceneuralnetworks} of component understanding in novel compositional context surroundings and mitigates the learning of spurious correlations in the training data~\cite{ye2024spuriouscorrelationsmachinelearning}. 
In addition, we show that using LCS to select challenging compositions of elusive components and novel contexts when augmenting data can decrease the Rademacher complexity~\cite{rademacher} term in the generalization bound, which provides an explanation for why LCS achieves better compositional generalization performance compared to purely random augmentation to some extent. 
(2) We extend the key ideas of our algorithms to the recently emerging and powerful LLMs' \textbf{I}n-\textbf{C}ontext \textbf{L}earning (ICL) scenarios~\cite{gpt3,dong2024surveyincontextlearning} (Section~\ref{sec:lcs_icl}). 
Unlike traditional learning paradigms, ICL only prepends a few exemplars (i.e., input-output pairs from an off-the-shelf demonstration pool, so-called ``demonstrations'') to the beginning of the input query and leverages these demonstrations to activate LLMs' ability to solve the query without updating LLMs' parameters.
However, recent works~\cite{drozdov2023compositional,an-etal-2023-context,gupta-etal-2023-coverage,gupta2024gistscore} have shown the deficiency of the ICL in compositional generalization.
To address this, we propose the LCS-ICL algorithm, which introduces additional compositional inductive bias by recombining existing demonstrations. This method enhances the exposure of the composition of elusive concepts and novel context surroundings by selecting the most informative recombined demonstrations according to the feedback from the LLMs.
(3) We enrich the experimental section with additional analysis experiments (Section~\ref{sec:exp_analysis}). For example, we present the experiment results on another challenging \textit{length} split of SCAN dataset and additional experiments using the Transformer~\cite{transformer} as the base model across all splits of the SCAN dataset in Table~\ref{tab:scan_exps}. We also include a new Figure (Figure~\ref{figure:fewshot}) to demonstrate the superiority of our CompSub data augmentation method in the low-data regimes of the GeoQuery dataset compared to other state-of-the-art data augmentation methods. Furthermore, we provide a comprehensive evaluation of our LCS-ICL algorithm using LLaMA2-13B~\cite{touvron2023llama2openfoundation} and LLaMA3-8B~\cite{dubey2024llama3herdmodels} in Table~\ref{tab:cofe_icl_prompt}, highlighting the effectiveness of our approach.

In summary, the main contributions of this paper are three-fold:
\begin{itemize}
    \item We introduce a new compositional data augmentation method, CompSub, which is the first to explore span-based compositional data augmentation, thus flexibly supporting to inject multi-grained compositional bias in to the training set. We theoretically analyze its effectiveness by proving that using CompSub is equivalent to imposing an implicit regularization term of learning the semantic invariance of individual components in different contexts on the optimization objective. Empirically we demonstrate its superiority on three benchmarks (SCAN, COGS and GeoQuery), with the improvement of at most $60.3\%$, $9.8\%$ and $1.2\%$ over previous state-of-the-art (SOTA) methods, respectively. 
    \item Based on CompSub, we introduce LCS as a differentiable data augmentation framework that first empowers difficulty-aware composition, being compatible with various down-stream language models. We theorectically analyze its effectiveness by proving that using LCS can decrease the Rademacher complexity term in the compositional generalization bound. We empirically demonstrate its superiority on three benchmarks (SCAN, COGS and GeoQuery), with the improvement of at most $66.5\%$, $10.3\%$ and $1.4\%$ over previous SOTA methods, respectively.
    \item We extend the key ideas of CompSub and LCS to ICL scenarios of LLMs, proposing LCS-ICL to enhance the few-shot compositional generalization capacity of state-of-the-art LLMs. We conduct comprehensive experiments on the COGS-QL dataset and show that LCS-ICL can exceed previous SOTA methods by at most $8.8\%$.
\end{itemize}
\section{Related Work}
\subsection{Compositional Generalization in Neural Language Modeling} A large body of literature pursues various ways of introducing compositional inductive bias into neural language models, in a bid to improve compositional generalization. 
The first category of studies, e.g., CGPS~\cite{primsub}, SyntAtt~\cite{synt_att}, GroupEqu~\cite{permutation-eq}, MultiGroupEqu~\cite{baltaji2024efficient} customizes neural architectures that promote lexical generalization via explicit disentanglement of  the meaning of tokens.
The second strand aims to align words or substructures in the input sequences with their counterparts in the output sequences by auxiliary tasks 
(e.g., 
IR-Transformer~\cite{IR-transf}), 
additional architectural modules 
(e.g., LexLearn~\cite{lexlearn}, Composed Layer Module~\cite{lin-etal-2023-learning} and LRF~\cite{zheng2024layer}),
as well as extra objectives imposed on attention layers (e.g., SpanAtt~\cite{span_attention}).
Third, the works of Meta-seq2seq~\cite{meta-seq2seq}, Comp-MAML~\cite{meta-comp}, and MET~\cite{mutual} resorts to the meta-learning paradigm 
to directly encourage compositional generalization of neural models. 
Last but not least,
compositional data augmentation that composes in-distribution data to accommodate out-of-distribution compositional sequences has been empirically demonstrated to enjoy not only the performance but also the model-agnostic benefits.
The explored principles for augmentation include exchangeability of tokens in the same context (e.g., GECA~\cite{geca}),  
token-level mixup~\cite{mixup} (e.g.,
SeqMix~\cite{seqmix}),
group-equivariance of 
language models~\cite{Basu2022EquiTuningGE,baltaji2024efficient}
by substituting training tokens (e.g., LexSym~\cite{lexsym}, Prim2PrimX~\cite{mutual}, ARCHER~\cite{cazzaro-etal-2024-align}) or subtrees (e.g., SUBS~\cite{subs}) 
with virtual or off-the-shelf tokens or subtrees.
Note that the aforementioned approaches guarantee the validity of composed sequences by following the widely accepted alignment practices in NLP, e.g., SpanTree~\cite{spanparse} and FastAlign~\cite{fastalign}.
Our work further pushes ahead with compositional data augmentation by (1) substituting spans, which offers more diverse and flexible generalization than substituting monotonous tokens or subtrees, and (2) endowing the augmentation strategy to be differentiable and learnable in an end-to-end manner, which dynamically adapts to the difficulty of down-stream neural sequence tasks. 

\subsection{Compositional Generalization in In-Context Learning of LLMs}
Recent works~\cite{qiu2022evaluatingimpactmodelscale,hosseini-etal-2022-compositional,chain_of_thought,drozdov2023compositional,an-etal-2023-context,gupta-etal-2023-coverage,gupta2024gistscore} also study the compositional generalization performance of LLMs in ICL scenarios.
Early works by Qiu et al.~\cite{qiu2022evaluatingimpactmodelscale} and Hosseini et al.~\cite{hosseini-etal-2022-compositional} comprehensively investigate the impact of the scale of LLMs and the number of demonstrations (i.e., \textit{shot number}) in ICL prompts on their compositional generalization performance, revealing a significant \textit{compositional generalization gap}~\cite{hosseini-etal-2022-compositional} in the state-of-the-art LLMs.
To address this gap, the first category of works~\cite{chain_of_thought,drozdov2023compositional} design ``Chain-of-Thought''-like ICL prompts to decompose complex queries into simpler sub-problems and then solve them step-by-step, thereby enhancing compositional generalization.
However, these methods require experts to carefully design the prompting templates and may suffer from poor generalization from the in-context demonstrations to unseen queries~\cite{zhou2023leasttomost}.
The second category of works~\cite{an-etal-2023-context,gupta-etal-2023-coverage,gupta2024gistscore} focus on retrieving appropriate exemplars from a pool of candidates as the ICL demonstrations. 
PrimCoverage~\cite{an-etal-2023-context} and BM25~\cite{bm25} aim to retrieve demonstrations that are similar to the query (covering parts of the query in the retrieved exemplars).
In contrast, CSR and BSR~\cite{gupta-etal-2023-coverage} emphasize the importance of selecting exemplars that, while less similar, contain more missing information in the query (i.e., more ``informative'') as ICL demonstrations.
GSR~\cite{gupta2024gistscore} trains an encoder for exemplar retrieval, which induces an attention masking bottleneck between exemplar inputs and outputs~\cite{mu2023learning}. By training with this bottleneck comprising a few gist tokens, the model is forced to store task-specific salient input information in the activations of those tokens, which GSR uses to retrieve the most informative ICL demonstrations.
Our work also falls into the second category of works. Different from existing works, we leverage our compositional data augmentation technique to inject additional compositional inductive bias into the ICL demonstrations and select demonstrations that are most helpful for compositional generalization based on the perplexity of LLMs themselves.
\section{Methodology}
\label{sec:method}
In this section, we present the details of our proposed algorithms. We first propose a novel compositional augmentation strategy, CompSub, which enables multi-grained composition of substantial substructures in the whole training set (Section~\ref{sec:spansub}). 
Over and above that, we introduce the LCS framework which empowers the learning of span substitution probabilities in CompSub in an end-to-end manner by maximizing the loss of neural language models, so as to outweigh those challenging compositions with elusive concepts and novel context surroundings (Section~\ref{sec:lcs}). 
Furthermore, we extend the key ideas of CompSub and LCS to the recently emerging in-context learning scenarios of pre-trained LLMs, proposing the LCS-ICL algorithm to enhance the few-shot compositional generalization of SOTA LLMs. (Section~\ref{sec:lcs_icl}).
\subsection{Component Substitution}
\label{sec:spansub}
\begin{figure}
\centering 
\includegraphics[width=0.47\textwidth]{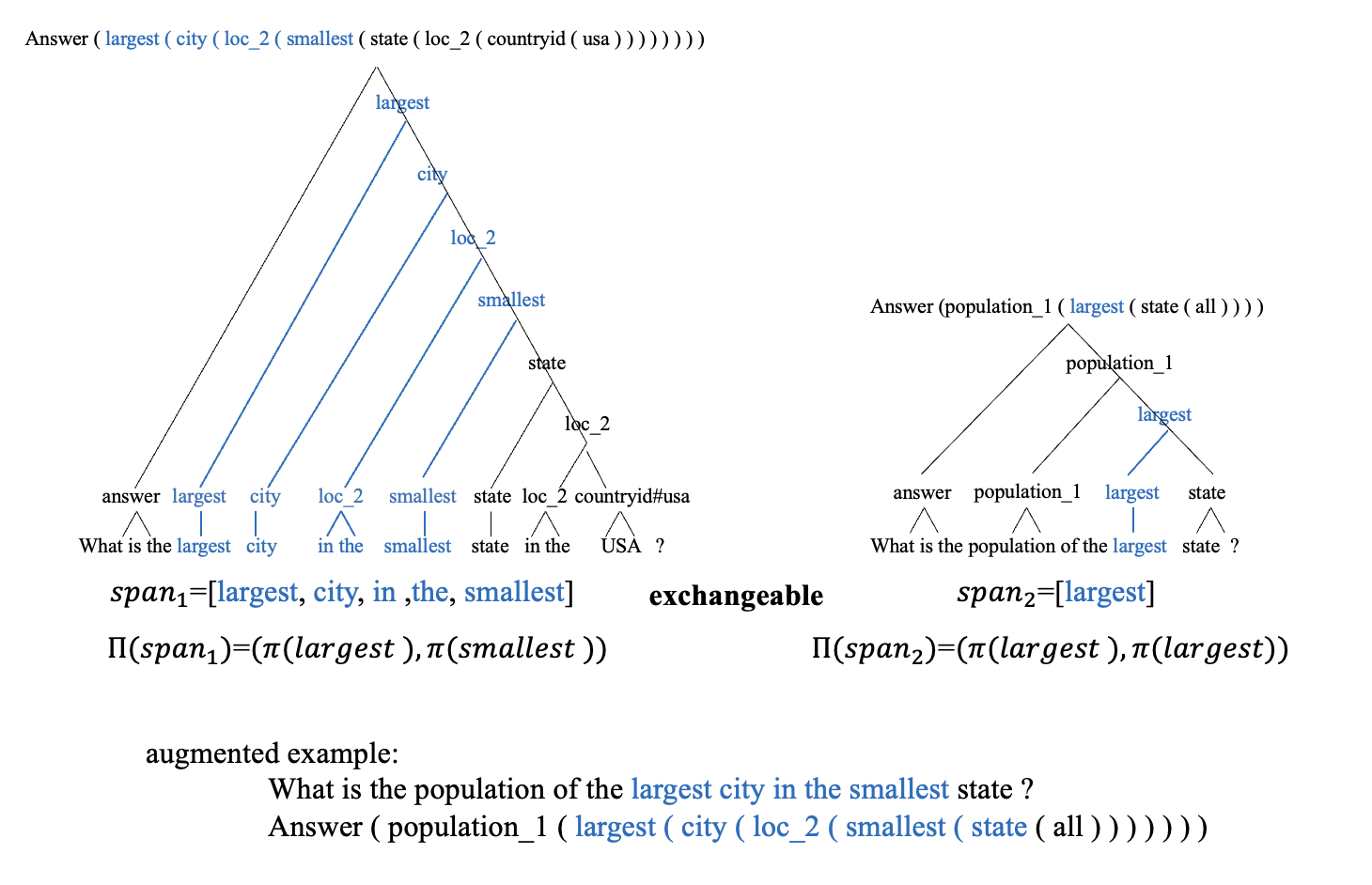}
\caption{An augmentation example by CompSub. 
CompSub substitutes a span ``largest'' with another span ``largest city in the smallest'', and augments a new question ``What is the population of the largest city in the smallest state?''.}
\label{figure:spansub}
\vspace{-0.1in}
\end{figure}
We propose CompSub to generate novel examples through exchanging multi-grained spans, which refer to consecutive fragments in input sequences, of the same equivalence class between training examples as shown in Fig.~\ref{figure:spansub}. 
Before proceeding to the details of CompSub, we first introduce two preprocessing prerequisites for CompSub, including extraction of span alignment and inference of the equivalence class of a word. 
On top of these, we present our substitution strategy that dictates the equivalence and exchangeability between spans.
\textbf{Preprocessing} The techniques of 
extracting span alignment from paired linguistic data and identifying syntactically equivalent words 
(e.g., Part-of-Speech tagging) have been well studied in the NLP community.
Following the practice in
a wealth of literature on compositional augmentation~\cite{lexsym, subs, mutual}, 
we also directly adapt the off-the-shelf techniques, which we introduce as below for self-contained purpose, to preprocess rather than delving into them. 
More details about the CompSub Algorithm (including the results of preprocessing for all the datasets) are available in the Supplementary Materials (Section \uppercase\expandafter{\romannumeral1}). \\
\textbf{Extraction of span alignment}
Span alignment refers to establish the correspondence between spans in the input sequence (e.g.,  ``largest city in the smallest'') and their counterparts (e.g.,   ``largest(city(loc\_2(smallest())))'') in the output sequence of a training example. 
For the SCAN dataset, we extract span alignment by extending SimpleAlign~\cite{lexlearn} that targets single words (e.g., \emph{jump $\rightarrow$ JUMP} \emph{right $\rightarrow$ TURN\_RIGHT}) to support alignment of consecutive fragments (e.g., \emph{jump right $\rightarrow$ TURN\_RIGHT JUMP}). 
As there always exists a deterministic function program~\cite{IR-transf,subs} that transforms the output sequence $y$ to a tree for COGS and GeoQuery, we resort to the intermediate representation~\cite{IR} of COGS from~\cite{IR-transf} and the span tree of GeoQuery from~\cite{spanparse} to map the input sequence $x$ to the tree form $T$, respectively.
The tree $T$, in such a way, serves as a bridge to align the input and output.\\
\textbf{Inference  of the equivalence class of a word}
\label{preprocess:cluster_token}
The aim is to infer the equivalence class of a word $w$, i.e., $\pi(w)$, according to the cluster it belongs to. Exemplar clusters include verbs and nouns. Fortunately, the COGS dataset has intrinsic clusters of words by their 
tree structure representations.
As for SCAN and GeoQuery, we follow~\cite{lexsym, mutual} to assign those words sharing the context into a single cluster. 
For example, the words of ``largest'' and ``smallest'' fall into the same cluster in Fig.~\ref{figure:spansub}.
\textbf{Substitution Strategy}
\begin{figure}
\centering 
\includegraphics[width=0.47\textwidth]{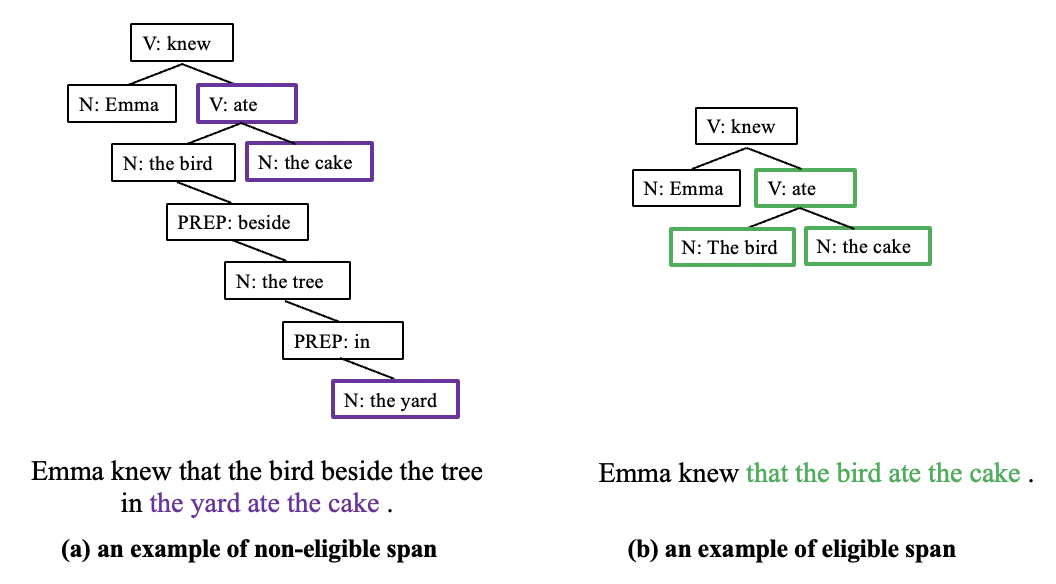}
\caption{Examples of non-eligible and eligible spans in COGS. (a) shows a non-eligible span which corresponds to an union set of disconnected fragments of the tree.}
\label{figure:eligible}
\vspace{-0.1in}
\end{figure}
The equivalence or exchangeability of spans, which a substitution strategy aims to establish, boils down to answering the following two questions: (1) what is an eligible span? (2) how to define the equivalence? First, given a consecutive span ${s = [w_p, w_{p+1}, ..., w_{p+k}]}$ where ${w_{p+i}} \ (0\leq i\leq k)$ represents a semantic unit (i.e., a word with semantic meaning), we define the span to be eligible if and only if it is semantically self-contained and unitary. Fig.~\ref{figure:eligible} shows a non-eligible span example ``the yard ate the cake'' which corresponds to an union set of two disconnected fragments of the tree and has an ambiguity (the subject of ``ate'' should be ``the bird'' rather than ``the yard''.). Such constraints imposed on eligible spans prevent substitutions with duplicate or missing parts. Due to page limit, we leave the formal mathematical definition of an \textbf{\textit{eligible}} span into the Supplementary Materials (Section \uppercase\expandafter{\romannumeral1}.C, Definition 1). 
Second, we formalize a heuristic rule to define the equivalence class of an eligible span $s$ as the combined equivalence classes of its first and last token, i.e.,

\vspace{-0.15in}
\small
\begin{align}
\Pi(s) \!=\! \Pi([w_p, w_{p+1}, ..., w_{p+k}]) \!=\! (\pi(w_p),\pi(w_{p+k})), 
\end{align}
\normalsize
where $\pi$ indicates the equivalence class of a single word as defined in Section \ref{preprocess:cluster_token}. 
By defining as above, it is legal to substitute a span ${s_1}$ with another span ${s_2}$ if and only if (1) both $s_1$ and $s_2$ are eligible and (2) $\Pi(s_1)=\Pi(s_2)$. Detailed pseudo codes of CompSub is also available (i.e., Algorithm.~\ref{algo:spansub}).

When dealing with tree structured tasks like GeoQuery and COGS, there are two special cases that need to be considered:
\begin{itemize}[leftmargin=*,noitemsep,topsep=0pt]
    \item ${s}\!=\![w_p]$ (e.g., ``largest'' in Fig.~\ref{figure:spansub}) degenerates to a single word: we specify that ${s}$ can only be substituted with another span $s'$ (either degenerated or undegenerated) with $\Pi(s')=[\pi(w_p),\pi(w_p)]$. 
    \item $s$ is a subtree with its root token $w_r$: we specify that ${s}$ can exchange with either another subtree $s'$ with $\Pi(s')=[\pi(w_r),\pi(w_{r})]$ or another span $s'$ with $\Pi(s')=[\pi(w_p),\pi(w_{p+k})]$).
\end{itemize}

\begin{algorithm}[t]
\label{algo:spansub}
	\caption{\textbf{CompSub}}
        \label{algo:spansub}
	\KwIn{Original dataset $\mathcal{D}$, the number of generated examples ${N}$, Span-Alignments extraction algorithm ${\mathcal{A}}$, Span-Classification function ${\Pi}$, Iterative Depth ${K}$.}
	\KwOut{Augmented dataset ${\mathcal{D}_{aug}}$.}  
	
        ${align}$, ${spans}$ $\leftarrow$ Run ${\mathcal{A}}$ on ${\mathcal{D}}$;  \\
        ${\mathcal{D}_{train}} \leftarrow\ \mathcal{D}$;\\
        \For{i $\leftarrow$ 1 to $K$}{
        ${\mathcal{D}_{aug}} \leftarrow\ \{\ \}$;\\
        
	\For{j $\leftarrow$ 1 to $N$}{
		   
	       Uniformly draw $d \in \mathcal{D}_{train}$ ;\\
        
        $({{x},{y}})\leftarrow d $;\\
       
        Uniformly draw span ${s}$ from ${inp}$;\\
        
		Uniformly draw span ${s'}\in \{ v| v\in {spans}, {\Pi(v)} = {\Pi(s)}\}$;\\
  
            ${x}_{aug} \leftarrow$ substitute ${s}$ with ${s'}$ in ${x}$;\\
            
            ${y}_{aug} \leftarrow$ substitute ${align(s)}$ with ${align(s')}$ in ${y}$;\\
            
            $d_{aug}\leftarrow ({x}_{aug},{y}_{aug}) $;\\
            
            ${\mathcal{D}_{aug}} \leftarrow {\mathcal{D}_{aug}} \cup \{d_{aug}\}$ \Comment{\textcolor{blue}{deduplication}}
	}
        $\mathcal{D}_{train} \leftarrow \mathcal{D}_{aug} \cup \mathcal{D}_{train}$;\\
        }
	\Return{${\mathcal{D}_{aug}}$}
\end{algorithm}

\subsection{Learning-based Component Substitution}
\label{sec:lcs}
\begin{figure}
\centering 
\includegraphics[width=0.47\textwidth]{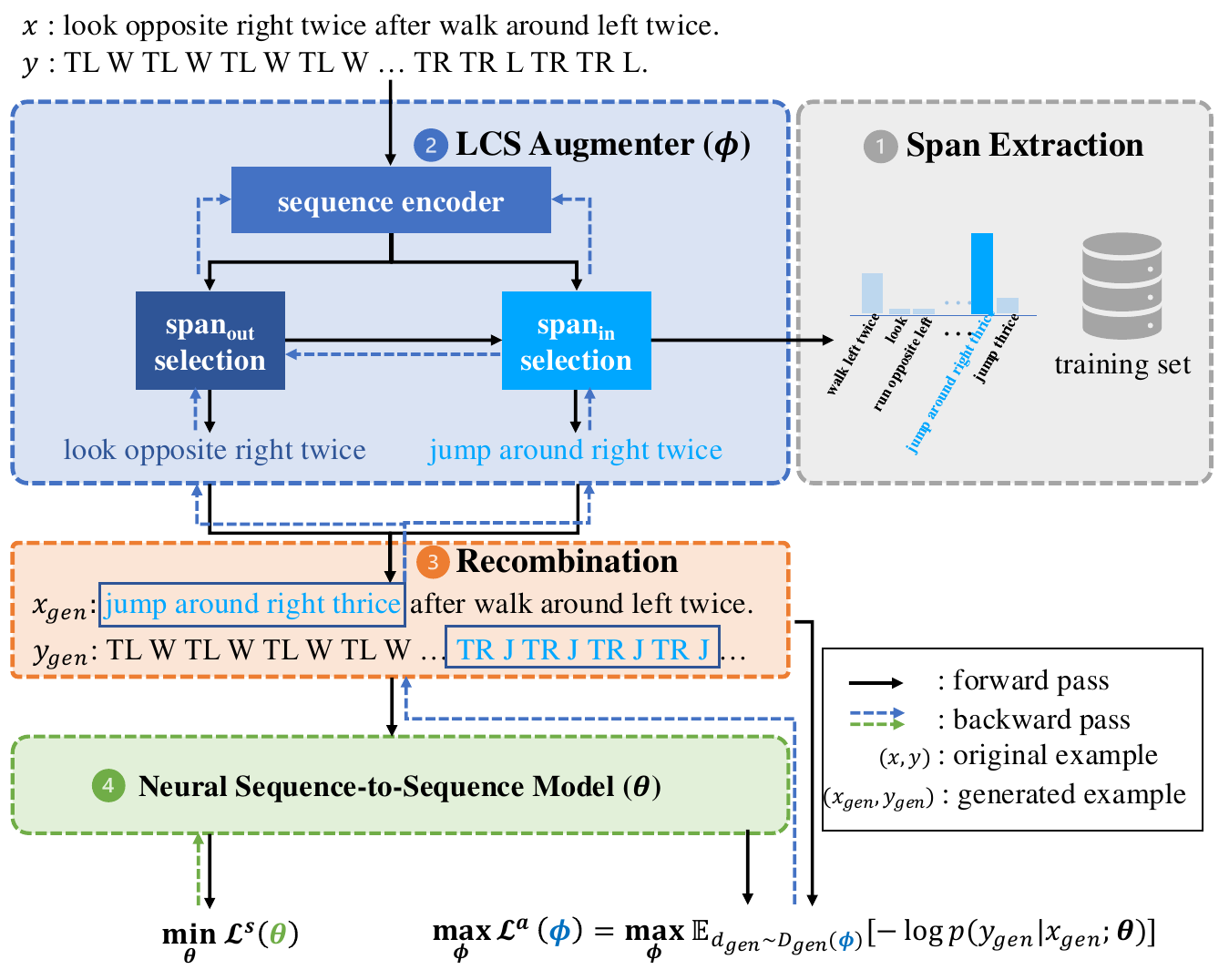}
\caption{Illustration of the LCS training framework. LCS training framework contains an upstream LCS augmentor and a downstream neural seq-to-seq model. Given an original training example $(x,y)$, the upstream LCS augmentor (parameter:$\phi$) predicts the probability distribution of the spans in $(x,y)$ to be substituted out and the probability distribution of the spans in the training set to be substituted in. Sampling the spans to be substituted out and substituted in from the above distributions, we augment the original training example to generate $(x_{gen},y_{gen})$ and send it into the down stream neural seq-to-seq model (parameter:$\theta$). In the parameter-update phase, we iteratively update $\phi$ by maximizing the loss of the downstream model and update $\theta$ by minimizing the loss of the downstream model.}
\label{figure:l2s2}
\end{figure}

Beyond the benefit of multi-grained compositional bias introduced by CompSub, the following three observations lead us to take a step further towards augmentation with attention on challenging spans.
(1) The 
distinct combinations for a linear number of distinct spans could be as many as the super-linear number~\cite{find_needles}. 
(2) The spans constitute both easy-to-comprehend and elusive ones, while oftentimes elusive ones are so rare that those combinations by them account for a very small portion.
(3) It is imperative to increase the percentage of these minority combinations to improve
the compositional generalization in a broad range of down-stream tasks.
Concretely,
we 
introduce an online and optimizable LCS framework consisting of a LCS augmenter that 
inherits the idea of span substitution with CompSub. 
More importantly, 
through maximizing the loss of down-stream neural sequence models, we learn span substitution probabilities in the upstreaming LCS augmenter to put high values on those chanllenging compositions of elusive spans and novel surroundings. The overview of the LCS framework is shown in Fig.~\ref{figure:l2s2}. 

\paragraph{Parameterizing the LCS Augmenter.}
\label{section:parameterization}

Given a training example $\bm{d}\!=\!{(x, y)}$, the objective of the LCS augmenter is to synthesize 
a new example ${\bm{d}_{gen}}\!=\!({x_{gen}}, {y_{gen}})$ via a sequence of~two actions ${\bm{a}\!=\!(a_{out},a_{in})}$: (1) $a_{out}$ which selects the
span ${s_{out}}$ to be swaped out from the span set $\mathcal{S}_1\!=\!\{s_1^i\}_{i=1}^u$ extracted from ${x}$\footnote{We can also identify spans in the ${y}$. This depends on the task type.}, and (2) 
$a_{in}$ which selects the span ${s_{in}}$ to be swapped in from the span set $\mathcal{S}_2\!=\!\{s_2^i\}_{i=1}^v$ extracted from the whole training dataset, following $a_{out}$.
Note that the preprocessing and span set extraction procedures are similar with Section~\ref{sec:spansub}, and $\mathcal{S}_1\!\subset\!\mathcal{S}_2$.
Once ${s_{out}}$ and ${s_{in}}$ are~selected, we have ${\bm{d}_{gen}}$ 
via recombination, i.e.,
\begin{itemize}[itemsep=0pt]
    \item ${x_{gen}}$ = ${x}$.replace(${s_{out}}$,${s_{in}}$),
    \item ${y_{gen}}$ = ${y}$.replace(${align(s_{out})}$,${align(s_{in})}$),
\end{itemize}
where replace($p,q$) denotes $p$ is replaced with $q$.

The probability of generating an ideal ${d_{gen}}$ based on ${d}$ 
is intuitively 
factorized as follows:
\begin{align}
  &p({\bm{d}_{gen}}|\bm{d};\bm{\phi})=p(\bm{a}|\bm{d};\bm{\phi})=p({(a_{out},a_{in})}|\bm{d};\bm{\phi}) \nonumber \\
 &=p({a_{out}}|\bm{d};\bm{\phi})\cdot p({a_{in}}|{a_{out}},\bm{d};\bm{\phi}) 
\end{align}
where $\bm{\phi}$ denotes the parameters of the LCS augmenter. In the following, we will detail how to model the two probabilities, during which we will introduce the 
the three parts that constitute $\bm{\phi}$.\\
\textbf{Parameterizing $p({a_{out}}|\bm{d};\bm{\phi})$ for selection of spans to be substituted out} 
Whether a span should be swapped out conditions on the equivalence class and the surroundings of the span, which are dictated by the representation of the span and that of the original training sequence ${x}$, respectively. To this end, we formulate the probability distribution $p({a_{out}}|\bm{d};\bm{\phi})$ over all $u$ candidate spans in $S_1$ as follows,
\begin{align}
\label{eq:out_distribution}
p({a_{out}}|\bm{d};\bm{\phi})={\tau(\mathcal{M}(\bm{\phi}_e(x),\bm{\phi}_o(\mathcal{S}_1)))},
\end{align}
where $\bm{\phi}_e$ as the first part of $\bm{\phi}$ represents the parameters of a sequence encoder ${\mathcal{R(\cdot)}}$, and $\bm{\phi}_o$ (the second part of $\bm{\phi}$)  denotes the embedding module for each candidate span in the span set $\mathcal{S}_1$. $\mathcal{M}(\cdot,\cdot)$ is a similarity function that measures the distance between two vectors. ${\tau}$ refers to the gumbel-softmax function~\cite{gumbel-softmax}, 
which guarantees sampling of the span with the largest probability, i.e., ${a_{out}^*}\sim p({a_{out}}|\bm{d};\bm{\phi})$, to be differentiable.
Implementation of the sampled action ${a_{out}^*}$ results in the selected span 
${s_{out}^*}$ 
to be substituted out.\\
\textbf{Parameterizing $p({a_{in}}|{a_{out}};\bm{d};\bm{\phi})$ for selection of spans to be substituted in}
The factors that govern the selection of a span to be swapped in from the whole span set $\mathcal{S}_2$ include the representations of (1) the span itself, (2) the input sentence ${x}$ for augmentation, and (3) the previously selected swap-out span ${s_{out}^*}$, so that those elusive spans that share the equivalence class with ${s_{out}^*}$ but contribute novel compositions via recombination with surroundings in $x$ are prioritized. 
Consequently, the probability distribution $p({a_{in}}|{a_{out}},\bm{d};\bm{\phi})$ over all $v$ candidate spans in $S_2$ follows,
\begin{align}
\label{eq:in_distribution}
 &\mathbf{c}=[\bm{\phi}_e(x);\bm{\phi}_{o}(s_{out}^*)]), \nonumber\\
 &p({a_{in}}|{a_{out}},\bm{d};\bm{\phi}) = \tau(\mathcal{M}(\bm{\phi}_{f}(\mathbf{c}),\bm{\phi}_i(\mathcal{S}_2))), 
\end{align}
where $\bm{\phi}_{f}$ and $\bm{\phi}_{i}$ altogether act as the third part of $\bm{\phi}$. Specifically, $\bm{\phi}_{i}$ is the embedding module for all spans in the span set $\mathcal{S}_2$ and $\bm{\phi}_{f}$ aligns the concatenated representation of the sentence and the swap-out span, i.e., $\bm{c}$, with $\bm{\phi}_i(\mathcal{S}_2)$ into the commensurable space.
Being consistent with the previous paragraph, we leverage the similarity function $\mathcal{M}(\cdot,\cdot)$ and the gumbel-softmax trick $\tau$ to sample ${a_{in}^*}\!\sim\! p({a_{in}}|{a_{out}^*},\bm{d};\bm{\phi})$.
It is noteworthy that we manually set the probability 
${a_{in}}\!\rightarrow\! {0}$ if ${\Pi(s_{in})\neq \Pi(s_{out}^*)}$ to excluse those potentially illegal synthesized examples.
The action ${a_{in}^*}$ finalizes 
the span ${s_{in}^*}$ 
to be substituted in.

\paragraph{Training Procedures for LCS.}
Training LCS boils down to two alternating procedures: first, the generated examples by the LCS augmenter pass forward to train the down-stream neural sequence-to-sequence model parameterized by $\bm{\theta}$; second, the performance of the neural sequence model serves as feedback to update the upstream augmenter parameterized by $\bm{\phi}=\{\bm{\phi}_{e},\bm{\phi}_{o},\bm{\phi}_{i},\bm{\phi}_{f}\}$. \\
\textbf{Training objective for the seq-to-seq model} 
The 
objective of training the 
seq-to-seq model is to minimize the expected 
negative log-likelihood of producing the output sequence ${y_{gen}}$ from the input one ${x_{gen}}$ conditioned on the its parameters $\bm{\theta}$, i.e.,

\small
\begin{align}
\label{obj:theta}
\min_{\bm{\theta}}\mathcal{L}^{s}(\bm{\theta}) &=   {\min_{\bm{\theta}}{\mathbb{E}_{\bm{d}_{gen}\sim \mathcal{D}_{gen}}[-\log p(y_{gen}|x_{gen};\bm{\theta})]}}\nonumber \\
   & \approx \min_{\bm{\theta}}-\frac{1}{NT}\sum_{n=1}^{N}\sum_{t=1}^T\log p(y_{gen}^{n,t}|x_{gen}^{n,t};\bm{\theta}).
\end{align}
\normalsize
We would highlight that the empirical estimation samples over not only $N$ examples but also $T$ sequences of actions for each example, thus avoiding the randomness and high variance induced by the gumbel softmax trick. Thus, $(x_{gen}^{n,t},y_{gen}^{n,t})$ denotes a generated example from the $n$-th original training example by following the $t$-th sampled action sequence $(a^{n,t}_{out},a^{n,t}_{in})$. $\mathcal{D}_{gen}$ represents the distribution of all generated samples by the augmenter.\\
\textbf{Training objective for the LCS augmenter}
Our main purpose is to encourage the upstream LCS augmenter
to outweigh those challenging compositions by the elusive spans and novel surroundings. 
To achieve this goal, we evaluate the difficulty of a newly composed example $\bm{d}_{gen}$ by 
the feedback from the down-stream seq-to-seq model, i.e.,
the negative log-likelihood of predicting it; the larger the negative log-likelihood is, the more challenging the generated example is. 
Intuitively, we solve the following optimization problem to train the LCS augmenter to maximize the difficulty of synthesized examples.

\small
\begin{align}
\label{obj:phi}
& \max_{\bm{\phi}}\mathcal{L}^{a}(\bm{\phi}) =   {\max_{\bm{\phi}}{\mathbb{E}_{\bm{d}_{gen}\sim \mathcal{D}_{gen}}[-\log p(y_{gen}|x_{gen};\bm{\theta})]}}\nonumber \\
   & \approx \max_{\bm{\phi}}-\frac{1}{NT}\sum_{n=1}^{N}\sum_{t=1}^T p(\bm{d}_{gen}^{n,t}|\bm{d}^{n,t};\bm{\phi})\log p(y_{gen}^{n,t}|x_{gen}^{n,t};\bm{\theta}),
\end{align}
\normalsize
where $p(\bm{d}_{gen}^{n,t}|\bm{d}^{n,t};\bm{\phi})$ refers to the gumbel softmax probability distribution of the $t$-th sampled action sequence $(a^{n,t}_{out},a^{n,t}_{in})$ that translates $\bm{d}^{n,t}$ into $\bm{d}_{gen}^{n,t}$. 
To keep the LCS augmenter timely posted of the training state of the neural seq-to-seq model, we alternatingly optimize these two parts. We present the pseudo codes for training LCS in Algorithm.~\ref{algo:l2s2}.

\begin{algorithm}[t]
	\caption{\textbf{Training LCS framework}}
        \label{algo:l2s2}
	\KwIn{Original dataset ${\mathcal{D}}$, \\LCS generator initialized parameters ${\phi_{0}}$, \\Seq-to-Seq Model initialized parameters ${\theta_{0}}$, \\Warm-up update number ${m}$,\\
    Sampled action number for each given example ${T}$. 
    }
	\KwOut{LCS generator parameters ${\phi_{f}}$, Seq-to-Seq Model parameters ${\theta_{f}}$.}  
	
        ${\theta \leftarrow \theta_{0}}$; ${\phi \leftarrow \phi_{0}}$
        
        \For{${step}\leftarrow$ 1 to m}{ 
            
            Sample ${\mathcal{B} \sim \mathcal{D}}$; 
            
            Optimize ${\theta}$ on ${\mathcal{B}}$ through Objective~\ref{obj:theta} 
        }
        
	\While{not converged}{
		  
            Sample ${\mathcal{B} \sim \mathcal{D}}$;
        
            \For{${t}\leftarrow$ 1 to T}{
            
            Sample ${\mathcal{B}_{gen,t} \sim p(\mathcal{B}_{gen}|\mathcal{B},\phi)}$;
            }
        
        Optimize ${\phi}$ on ${\{B_{gen,t}\}_{t=1}^T}$ through Objective~\ref{obj:phi} 
        
        Sample ${\mathcal{B} \sim \mathcal{D}}$;
        
        Sample ${\mathcal{B}_{gen} \sim p(\mathcal{B}_{gen}|\mathcal{B},\phi)}$;
        
        Optimize ${\theta}$ on ${B_{gen}}$ through Objective~\ref{obj:theta} 
	}
	\Return{${\phi,\theta}$}
\end{algorithm}

\subsection{Generalize the methods to In-Context Learning Scenarios}
\label{sec:lcs_icl}

\begin{figure*}
\centering 
\includegraphics[width=0.95\textwidth]{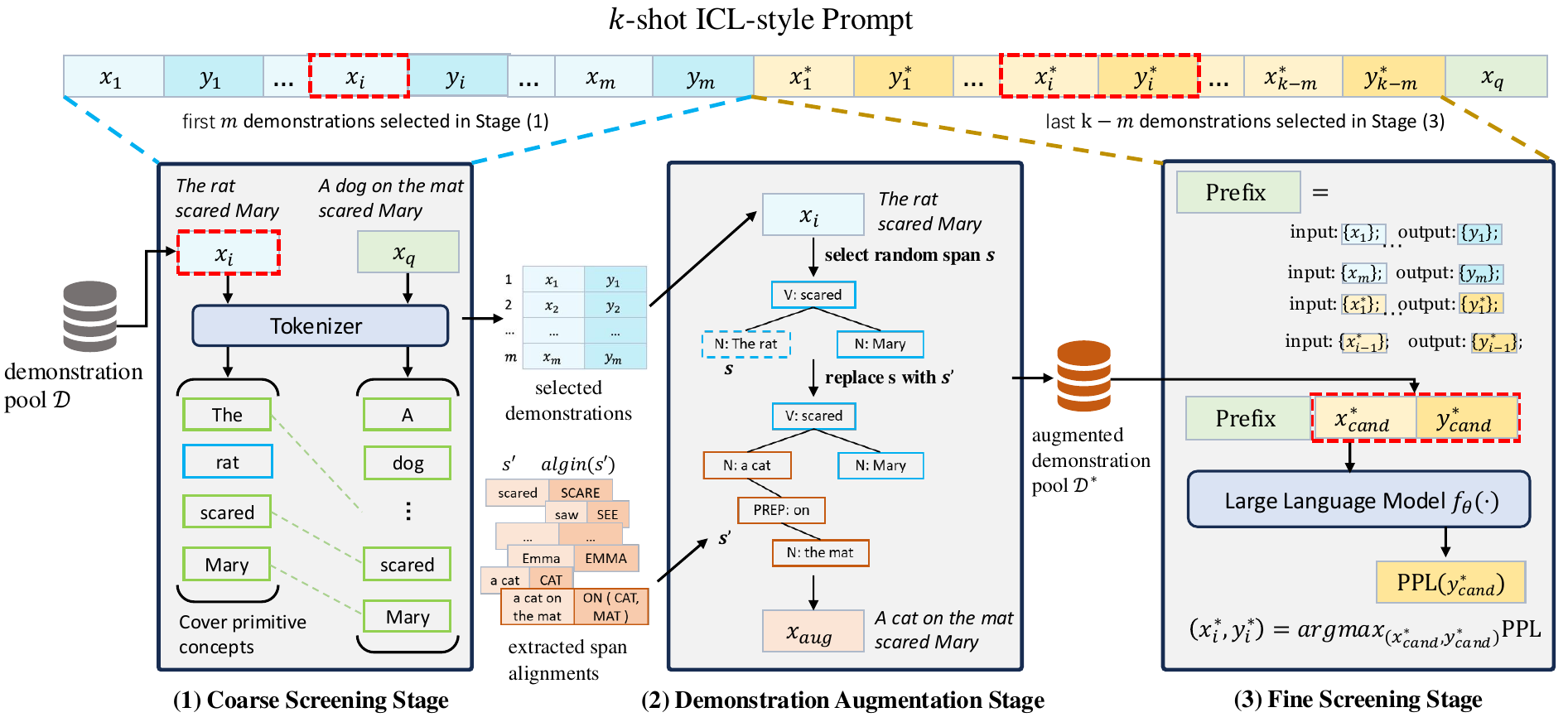}
\caption{The figure illustrates the workflow of the LCS-ICL algorithm when constructing a $k$-shot ICL-style prompt. The whole workflow mainly contains three stage. (1) Coarse Screening Stage: Select $m\approx\lceil k/2\rceil$ examples from $\mathcal{D}$: $\{(x_i,y_i)\}_{i=1}^{m}$ to guarantee that as many primitive concepts in the query $x_q$ are covered in $\{x_i\}_{i=1}^{m}$ as possible. (2) Demonstration Augmentation Stage: Introduce additional compositional inductive bias by running CompSub on $\{(x_i,y_i)\}_{i=1}^{m}$ to get an augmented demonstration pool $\mathcal{D^*}=\text{CompSub}(\{(x_i,y_i)\}_{i=1}^{m})$. (3) Fine Screening Stage: Successively retrieve the rest $n = k-m$ demonstrations from $\mathcal{D^*}$ with the policy of choosing the candidate demonstration that the model are difficult to handle (with the highest perplexity score) by in-context learning from currently selected demonstrations.}
\label{figure:lcs_icl}
\vspace{-0.1in}
\end{figure*}

Large language models (LLMs) have garnered significant attention primarily due to their effectiveness as "few-shot learners"~\cite{gpt3} and their ability to be adapted to various downstream tasks through the \textbf{I}n-\textbf{C}ontext \textbf{L}earning (ICL) paradigm~\cite{dong2024surveyincontextlearning}.
ICL typically involves selecting few-shot (e.g., $k$-shot) exemplars, comprising both inputs $\{x_i\}_{i=1}^k$ and outputs $\{y_i\}_{i=1}^k$ (referred to as "few-shot \textbf{demonstrations}"), from the available demonstration pool $\mathcal{D}$ (i.e., the training data in the classical learning paradigm) and attaching these demonstrations $\{(x_i,y_i)\}_{i=1}^k$ immediately before the real query input $x_{q}$ to form the prompt for the LLM.
In realistic applications of ICL, although $x_{q}$ may not directly appear in $\mathcal{D}$, it is often a recombination of components observed in $\mathcal{D}$~\cite{drozdov2023compositional,gupta-etal-2023-coverage}.
Hence, it is important to investigate and enhance the capacity of LLMs to grasp the meaning of individual components from the very limited demonstrations and generalize the acquired knowledge to the recombined queries (i.e., the compositional generalization under ICL).
We formally define the compositional generalization under the $k$-shot ICL of LLMs as follows: First we have a demonstration pool $\mathcal{D}$ and a testing set $\mathcal{D}_{test}$ in which each instance $(x_{q},y_{q})$ satisfies that $(x_{q},y_{q})\notin \mathcal{D}$ and $(x_{q},y_{q})$ can be derived through recombining the components that appear in the instances in $\mathcal{D}$. For each testing case $(x_{q},y_{q})\in \mathcal{D}_{test}$, we prepend $k$ demonstrations $(x_i,y_i)_{i=1}^k \in\mathcal{D}$ before $x_{q}$ to form a $k$-shot ICL prompt $\text{Prompt}(x_{q})$ and then use it to query the LLM $f_\theta(\cdot)$ (i.e., the predicted output $\hat{y_{q}}=f_\theta(\text{Prompt}(x_{q}))$). The compositional generalization performance under the $k$-shot ICL of LLMs is defined as the average prediction accuracy: $\frac{1}{|\mathcal{D}_{test}|}\sum_{(x_{q},y_{q})\in \mathcal{D}_{test}}\mathbb{I}(f_\theta(\text{Prompt}(x_{q})),y_{q})$.

The strategy of selecting the demonstrations for the given query is the key to the compositional generalization performance of the in-context learning.
Recent studies show that randomly sampling demonstrations from the demonstration pool results in poor compositional generalization~\cite{an-etal-2023-context,gupta-etal-2023-coverage}.
Inspired by previous theoretical insights on ICL~\cite{akyürek2023learningalgorithmincontextlearning,vonoswald2023transformerslearnincontextgradient}, \textit{LLMs implicitly perform parameter update via gradient descent on the few-shot in-context demonstrations}. Given that the number of demonstrations in the ICL prompt is typically much smaller than the entire demonstration set~\cite{liu2023lostmiddlelanguagemodels,li2024longcontextllmsstrugglelong}, adaptively determining appropriate few-shot demonstrations for a specific query significantly impacts the compositional generalization performance of the ICL in LLMs~\cite{an-etal-2023-context,gupta-etal-2023-coverage}.

To this end, we extend the main idea of the LCS method to the ICL scenarios, resulting in the LCS-ICL algorithm (the workflow of LCS-ICL is depicted in Figure~\ref{figure:lcs_icl}).
Our LCS-ICL algorithm incorporates two key ideas: one is recombining existing demonstrations to \textbf{\textit{introduce additional compositional inductive bias}} in the in-context learning stage; another is selecting the recombined demonstrations according to the feedback from the LLM (the most difficult demonstrations) to \textbf{\textit{enhance the exposure of the composition of elusive concepts and novel context surroundings}}.

Specifically, supposing that we are going to select $k$ demonstrations $\{(x_i,y_i)\}_{i=1}^k$ from the demonstration pool $\mathcal{D}$ for the current query $x_{q}$, we present the workflow of the LCS-ICL algorithm as the following three stages:
(1) \textbf{Coarse Screening Stage}: Select $m=\lceil k/2\rceil$ examples from $\mathcal{D}$: $\{(x_i,y_i)\}_{i=1}^{m}$ to guarantee that all primitive concepts (e.g., the query “The child on a table burned the pizza beside a stage" contains the concepts of $\{\text{“child", “on", “table", “burned", ..., “stage"}\}$) in the query $x_q$ are covered in $\{x_i\}_{i=1}^{m}$. $\{(x_i,y_i)\}_{i=1}^{m}$ are selected as the first $m$ demonstrations for the ICL prompt. Note that if $\lceil k/2\rceil$ demonstrations are insufficient to fully cover all primitive concepts in $x_q$, we select $m>\lceil k/2\rceil$ demonstrations until full coverage is achieved ($m\leq n$). We use the greedy algorithm for the Set-Cover problem to implement the first stage.
(2) \textbf{Demonstration Augmentation Stage}: Introduce additional compositional inductive bias by running CompSub (Algorithm~\ref{algo:spansub}) on $\{(x_i,y_i)\}_{i=1}^{m}$ to get an augmented demonstration pool $\mathcal{D^*}=\text{CompSub}(\{(x_i,y_i)\}_{i=1}^{m})$, where all of the demonstrations can be derived by recombining demonstrations in $(x_i,y_i)_{i=1}^m$.
(3) \textbf{Fine Screening Stage}: Successively retrieve the remaining $n = k-m$ demonstrations from $\mathcal{D^*}$ by selecting the examples that the model finds most challenging. When retrieving the $i\ (1\leq i\leq n)$ -th demonstration, we concatenate all previously selected demonstrations (including $m$ demonstrations $\{(x_i,y_i)\}_{i=1}^{m}$ from the coarse screening stage and $i-1$ demonstrations $\{(x^*_j,y_j^*)\}_{j=1}^{i-1}$ already retrieved from $\mathcal{D^*}$ in the fine screening stage) together and append a candidate demonstration $(x^*_{cand},y^*_{cand})\in \mathcal{D}^*/\{inp_j^*,out_j^*\}_{j=1}^{i-1}$ (from $\mathcal{D^*}$) after them to generate a ICL-style prompt:
\begin{align*}
\text{Prompt}=&\text{input:}\{x_1\};\text{output:}\{y_1\};\\
            &\text{input:}\{x_2\};\text{output:}\{y_2\};\\
            & ... \\
            &\text{input:}\{x_m\};\text{output:}\{x_m\};\\
            &\text{input:}\{x_1^*\};\text{output:}\{y_1^*\};\\
            & ... \\
            &\text{input:}\{x_{i-1}^*\};\text{output:}\{y_{i-1}^*\};\\
            &\text{input:}\{x_{cand}^*\};\text{output:}\{y_{cand}^*\}\\
\end{align*}         
We input the $\text{Prompt}$ into the large language model and calculate the Perplexity score\footnote{\url{https://en.wikipedia.org/wiki/Perplexity}} (PPL) of the $y_{cand}^*$ part (We define this value as the LCS-Score of this candidate demonstration). We traverse all possible candidates and select the one with the highest LCS-Score as the selected demonstration $\{x_i^*,y_i^*\}$. We then proceed to retrieve the next demonstration $\{x_{i+1}^*,y_{i+1}^*\}$ from the remaining pool $\mathcal{D^*}/\{x_j^*,y_j^*\}_{j=1}^i$.
In real applications, we typically select the demonstration from a random sampled subset $\mathcal{S}\subseteq \mathcal{D}^*$ (e.g., $|S|=100$) to speed up the running time.
The pseudo codes for the LCS-ICL algorithm is provided in the Supplementary Materials (Section \uppercase\expandafter{\romannumeral4}).
\section{Theoretical Insights}
\label{sec:th}
In this section, we aim to provide theoretical insights into the algorithms we introduced in Section~\ref{sec:method}. We demonstrate that: \emph{from the perspective of optimization}, our proposed compositional data augmentation strategy, CompSub (Algorithm~\ref{algo:spansub}), serves as an implicit regularization term. 
It characterizes the semantic invariance of specific components that, when combined with different context surroundings, make up the input text data. 
Learning this semantic invariance helps the language model disentangle the meaning of different individual components, thereby mitigating the spurious correlation~\cite{arjovsky2020invariantriskminimization, 10430110, li2024invariantrepresentationdecouplingstyle} between the target component and its context surroundings that co-occur frequently in the training set in the end-to-end training manner; 
\emph{from the perspective of generalization risk}, our learning-based compositional data augmentation framework (Algorithm~\ref{algo:l2s2}, after fully leveraging CompSub to extend the training data distribution) can be interpreted as automatically generating the most challenging compositional examples. 
This process reduces the Rademacher complexity~\cite{rademacher} and consequently, tightens the upper bound of the generalization risk. 
Due to the page limitations, we present most of the proof details in the Supplementary Materials (Section \uppercase\expandafter{\romannumeral2}).


Following the analysis framework of Chen et al.\cite{group_da_theory}, we formulate all possible component exchange operations (in the following discussion, we call each component exchange operation a `transform') as a permutation group~\cite{passman2012permutation} $\mathcal{G}$. Each permutation element $g\in\mathcal{G}$ naturally represents a set of component substituting operations (e.g., “jump left”$\rightarrow$“run around right”; “look right”$\rightarrow$“walk opposite left”; ...), where we apply these operations to compositionally augment the original training data.
For a given permutation element $g$ and an original data point $x\in\mathcal{X}$, we denote that $g\circ x$ refers to apply the corresponding component substitution operations to $x$. In alignment with the previous theoretical analysis~\cite{JMLR:v25:22-1312, pmlr-v139-elesedy21a} on data augmentation, we make the following assumption.
\begin{myAssume}
\label{lemma:orbit_average}
\textbf{(Unbiased on average)} Let $x$ represent a sample from the training set $D$, $f_\theta$ represent the neural network mapping function, and $g$ represent a transform operation sampled from the group $\mathcal{G}$. We assume that: $\mathbb{E}_{g\in\mathcal{G}}[f_\theta(g\circ x)] = f_\theta(x)$.
\end{myAssume}

Considering the example $(x,y)$ ($x$ refers to the input and $y$ refers to the output) in the training set, the input sequence $x$ could be divided into two parts $x_*$ (the key component we want to disentangle from the whole sequence) and $x_s$ (the surrounding context of $x_*$ in $x$): $x=x_* \oplus x_s$. Correspondingly, the output sequence $y$ could be divided into $y_*$ and $y_s$ (the counterparts of $x_*$ and $x_s$): $y=y_*\oplus y_s$. The operator $\oplus$ denotes the `composition' of a component and its surrounding context.
By way of example, in the SCAN dataset~\cite{scan}: $x=$``\textcolor{black}{jump around right twice} \textcolor{black}{and walk left}”, $y=$``\textcolor{black}{TR J TR J TR J TR J TR J TR J TR J TR J} \textcolor{black}{TL W}”. Supposing we focus on the component ``jump around right twice”, $x^*=$``\textcolor{black}{jump around right twice}”, $x_s=$``\textcolor{black}{and walk left}”, correspondingly $y^*=$``\textcolor{black}{TR J TR J TR J TR J TR J TR J TR J TR J}”, $y_s=$``\textcolor{black}{TL W}”.

The objective of modeling the sequence-to-sequence probability of a language model $\theta$ is $p_\theta(y|x)=p_\theta(y_s\oplus y_*|x_s\oplus x_*)$. 
Note that $y_s$ can be splitted into two parts $y_{s}^{1}$ and $y_{s}^2$. $y_{s}^1$ refers to the surrounding context that is ahead of $y_*$ and $y_{s}^2$ refers to the surrounding context that is behind of $y_*$.
We have the following probability decomposition: $p_\theta(y|x)=p_\theta(y_s\oplus y_*|x_s\oplus x_*)=p_\theta(y_{s}^1|x_s\oplus x_*)p_\theta(y_*|y_s^1,x_s\oplus x_*)p_\theta(y_s^2|y_*,y_s^1,x_s\oplus x_*)$.
In this section, we mainly consider the errors that frequently occur when predicting the key component $y_*$: $p_\theta(y_*|y_s^1,x_s\oplus x_*)$.
In the following paragraphs, for the simplicity, we directly use the notation $y_s$ to represent $y_s^1$ and we focus on the probability $p_\theta(y_*|y_s,x_s\oplus x_*)$.
\begin{myTheo}
\label{lab:theo1}
Let $h$ denote the negative likelihood loss function: $h(p) =-log(p)$.
We have the following inequality:
$
\mathbb{E}_{x_s,y_s}[h(p_\theta(y_*|y_s,x_s\oplus x_*)) + 2\sum_{g\in\mathcal{G}} \lVert p_\theta(y_*|g \circ y_s, (g\circ x_s)\oplus x_*)-p_\theta(y_*|y_s,x_s\oplus x_*)\rVert_2^2] \leq
\mathbb{E}_{x_s,y_s}\mathbb{E}_{g}[p_\theta(y_*|g \circ y_s,(g\circ x_s) \oplus x_*)]
$,
where the expression on the left hand side represents the original seq-to-seq loss function $\mathbb{E}_{x_s,y_s}[h(p_\theta(y_*|y_s,x_s\oplus x_*))$ with an additional regularization term $\sum_{g\in\mathcal{G}} \lVert p_\theta(y_*|g \circ y_s, (g\circ x_s)\oplus x_*)-p_\theta(y_*|y_s,x_s\oplus x_*)\rVert_2^2]$ and the expression on the right hand side represents the seq-to-seq loss function expected on the augmented dataset $\mathbb{E}_{x_s,y_s}\mathbb{E}_{g}[p_\theta(y_*|g \circ y_s,(g\circ x_s)\oplus x_*)]$.
\end{myTheo}

When we optimize $\theta$ with gradient descent based methods, we force the compositional augmentation loss $\mathbb{E}_{g,x_s,y_s}[h(p_\theta(y_*|g \circ y_s,(g\circ x_s)\oplus x_*))]$ to approach $0$, which meanwhile turns out to force right-hand side (RHS) to approach $0$.
Theorem~\ref{lab:theo1} shows that, based on the original loss term, our compositional data augmentation strategy implements an implicit regularization term $\sum_{g\in\mathcal{G}} \lVert p_\theta(y_*|g \circ y_s,(g\circ x_s)\oplus x_*)-p_\theta(y_*|y_s,x_s\oplus x_*)\rVert_2^2$. The meaning of this term is to encourage the model to predict the same $y_*$ given a specific $x_*$ and diverse surrounding contexts, i.e., to disentangle the meaning of the individual semantic component $x_*$ from the entire input $x$. 
\begin{myCorollary}
In the ideal conditions, where the training loss converges to zero and our compositional data augmentation fully substitutes the surrounding context $(x_s,y_s)$ for every $(x_*, y_*)$, we can disentangle the language modeling probability: $p_\theta(y_s\oplus y_* | x_s\oplus x_*) = p_\theta(y_s|x_s)\cdot p_\theta(y_*|x_*)$. 
\end{myCorollary}


Such disentanglement, which is widely pursued in the compositional generalization literature~\cite{primsub,samir-silfverberg-2023-understanding}, helps to learn the invariance~\cite{arjovsky2020invariantriskminimization,lyle2020benefitsinvarianceneuralnetworks} and eliminate spurious correlation~\cite{10430110,ye2024spuriouscorrelationsmachinelearning} in end-to-end language modeling and compositional generalization.

Denote the augmented training data distribution as $\mathcal{X}_{aug}$ and the unseen compositional testing data distribution as $\mathcal{X}_{comp}$. Let $\mathcal{S} = \{(x_i,y_i)\}_{i=1}^n$ be a sample set drawn independently identically distributed from $\mathcal{X}_{aug}$, $\mathcal{H}$ be a model class and $l:\mapsto[0,1]$ be a loss function. For a $h\in\mathcal{H}$, we define the empirical risk as $R_{\mathcal{S}}(h)=\frac{1}{n}\sum_{i=1}^{n}l(h(x_i),y_i)$, the expected risk $R_{aug}(h)=\mathbb{E}_{(x,y)\sim\mathcal{X}_{aug}}[l(h(x),y)]$ and the expected risk for the compositional generalization $R_{comp}(h)=\mathbb{E}_{(x,y)\sim\mathcal{X}_{comp}}[l(h(x),y)]$.
We have $R_{comp}(h)=R_{aug}(h) + (R_{comp}(h)-R_{aug}(h))$, where the second term $(R_{comp}(h)-R_{aug}(h))$ is upper bounded by $L\mathbb{W}_1(\mathcal{X}_{aug},\mathcal{X}_{comp})$~\cite{fu2024generaltheorycompositionalgeneralization}. $L$ is the lipschitz constant of $h$ and $\mathbb{W}_1$ denotes the 1-Wasserstein distance~\cite{10.5555/3540261.3541722}. This value of the expression $L\mathbb{W}_1(\mathcal{X}_{aug},\mathcal{X}_{comp})$ depends on the inherent ability of the data augmentation scheme: in comparison with previous compositional data augmentation scheme (e.g., SUBS~\cite{subs} and LexSym~\cite{lexsym}), CompSub can achieve a relatively low 1-Wasserstein distance between the augmented training data distribution and the true compositional testing data distribution (we present empirical results in Supplementary Materials Section \uppercase\expandafter{\romannumeral4}). 
Then we focus on analyzing the first term $R_{aug}(h)$.\\
\textbf{(Rademacher Complexity)} The empirical rademacher complexity of $\mathcal{M}$ ($\mathcal{M}$ is interpreted as the family of loss function $l$ associated with the model class $\mathcal{H}$ mapping from the data distribution $\mathcal{X}$ to $\mathbb{R}$) is defined as: $\mathcal{R}_\mathcal{S}(\mathcal{M})=\mathbb{\mathbf{\sigma}}[sup_{m\in\mathcal{M}}\frac{1}{n}\sum_{i=1}^{n}\sigma_i m(x_i,y_i)]$.
\begin{myTheo}
Given a substitution operation group $\mathcal{G}$, define $\hat{l}$ as $\hat{l}(x_i,y_i)=\max_{g\in\mathcal{G}}(l(h(g\circ x), g\circ y))$. We have that the expected risk for the compositional generalization $R_{comp}(h)$ is upper bounded by $R_\mathcal{S}(\hat{l}\circ h) + 2\mathcal{R}_\mathcal{S}(\hat{l}\circ h) + 3\sqrt{\frac{\log\frac{2}{\delta}}{2n}} + L\mathbb{W}_1(\mathcal{X}_{aug},\mathcal{X}_{comp})$.
\end{myTheo}


As in the previous discussion, the distribution distance is determined by the inherent ability of the compositional augmentation scheme. Hence we focus on the rademacher complexity term $\mathcal{R}_\mathcal{S}(\hat{l}\circ h)$. As shown in Zhu et al.~\cite{NEURIPS2021_2287c6b8}, $\mathcal{R}_\mathcal{S}(\hat{l}\circ h)$ is upper bounded by the original complexity $\mathcal{R}_\mathcal{S}(h)$. In the LCS training framework (Algorithm~\ref{algo:l2s2}), given an original case $(x,y)\in\mathcal{D}_{aug}$ we train a neural network to select components to substitute out and substitute in, generating the augment example $(x',y')$ that has the largest loss value. Ideally, the generated example $(x',y')=(g\circ x,g\circ y)$, where $g=\text{argmax}_{g}l(h(g\circ x),g\circ y)$. Hence we demonstrate that with the LCS algorithm, the Rademacher complexity will be decreased. Meanwhile, we use the generated examples $(x',y')$ to train the sequence model by minimizing the empirical risk $R_\mathcal{S}(\hat{l}\circ h)$ (to near zero). These factors together guarantee that the generalization bound $R_\mathcal{S}(\hat{l}\circ h) + 2\mathcal{R}_\mathcal{S}(\hat{l}\circ h) + 3\sqrt{\frac{\log\frac{2}{\delta}}{2n}} + L\mathbb{W}_1(\mathcal{X}_{aug},\mathcal{X}_{comp})$ is tighter compared with the naive training algorithm (training with the $l$ objective).

\section{Experiment}
In this section, we present our empirical results for the methods proposed in Section~\ref{sec:method}. Specifically, in Section~\ref{sec:exp_dataset}, we introduce the datasets and the corresponding evaluation metrics used to assess our methods. 
In Section~\ref{sec:exp_baseline}, we briefly review the representative baseline methods to demonstrate the superiority of our methods. 
In Section~\ref{sec:exp_results}, we present and describe the main experiments results.
Finally, in Section~\ref{sec:exp_analysis}, we provide a detailed analysis of the empirical results, including a breakdown of the performance, ablation studies, and additional insights.
\subsection{Datasets and Evaluation Metric}
\label{sec:exp_dataset}
We evaluate our proposed methods on the following four popular and representative datasets (including ten sub-tasks in total) which are specifically designed for assessing the compositional generalization capacity of neural language models. The datasets adopted in the paper are diverse for (1) all kinds of lexical-level generalization testing cases~\cite{primsub,scan,cogs,an-etal-2023-context} and structural-level generalization testing cases~\cite{MCD, cogs, geoquery, an-etal-2023-context} are covered in these datasets, making the evaluation holistic and (2) there are not only synthetic evaluations deliberately designed for diverse categories of compositional generalization but also non-synthetic ones additionally requiring capabilities of neural models in handling natural language variations~\cite{nqg}. In the following paragraphs, we introduce these datasets and tasks one-by-one. 
\\
\textbf{SCAN} Initially introduced by Lake and Baroni~\cite{scan}, SCAN contains a large set of synthetic paired sequences whose input is a sequence of navigation commands in natural language and output is the corresponding action sequence. Following previous works~\cite{geca,compibt,lexlearn,mutual,lexsym}, we evaluate our methods on the two splits of SCAN: \textit{\textbf{jump}} split (designed to evaluate a novel combination of a seen primitive, i.e., jump, and other seen surroundings) and \textit{\textbf{around right}} split (designed to evaluate a novel compositional rule). In particular, we also consider two more complex and challenging scenarios: \textit{\textbf{length}} split (designed to evaluate longer sequences when trained on shorter ones) and \textit{\textbf{Maximum Compound Divergence (MCD)}} splits of SCAN established in Keysers et al.~\cite{MCD}, which distinguish the compound distributions of the training and the testing set as sharply as possible.\\
\textbf{COGS} Another synthetic COGS dataset~\cite{cogs} contains 24,155 pairs of English sentences and their corresponding logical forms (with the Lambda calculus). COGS contains a variety of systematic linguistic abstractions (e.g., active $\rightarrow$ passive, nominative → accusative and transtive verbs $\rightarrow$ intranstive verbs), reflecting compositionality of natural utterance. It is noteworthy that COGS with its testing data categorized into 21 classes by the compositional generalization type supports fine-grained evaluations.\\
\textbf{GeoQuery} The non-synthetic dataset of GeoQeury~\cite{geoquery} collects 880 anthropogenic questions regarding the US geography (e.g., "what states does the mississippi run through?") paired with their corresponding database query statements (e.g., "answer ( state ( traverse$\_$1 ( riverid ( mississippi ) ) ) )"). Following ~\cite{IR,subs}, we also adopt the FunQl formalism of GeoQuery introduced by~\cite{funql} and evaluate our methods on the compositional template split (query split) from Finegan-Dollak et al.~\cite{template-split} where the output query statement templates of the training and testing set are disjoint and the i.i.d. split (question split) where training set and testing set are randomly separated from the whole dataset.\\
\textbf{COGS-QL} COGS-QL~\cite{an2023does, an-etal-2023-context} is a variant of COGS~\cite{cogs} which can be used for accessing large language models' compositional generalization capacity in the \textit{\textbf{In-Context Learning}} (ICL) scenarios. 
The traditional COGS task might be hard to be fully resolved by current open-sourced LLMs~\cite{touvron2023llama2openfoundation,dubey2024llama3herdmodels} due to its complex output form.
Different with COGS, COGS-QL reconstruct the output sequences from the Lambda calculus form to the FunQL~\cite{funql} form (which is much simpler), making the generalization task more focus on the compositional skills themselves. For experiments with LLaMA2-13B, we randomly sample 200 examples from the COGS-QL dataset to test the performance. For experiments with LLaMA3-8B, we randomly sample 1,000 examples from the COGS-QL dataset to test the performance.\\
\textbf{Evaluation Metric} Following the convention of previous works~\cite{scan,cogs,template-split,an2023does}, we adopt the evaluation metric of \textbf{\textit{exact-match accuracy}} in all of our experiments.
\subsection{Baselines and Base Models}
\label{sec:exp_baseline}
We compare our proposed approaches, including CompSub and LCS (for both training-based paradigm and ICL paradigm), with following prior state-of-the-art baseline approaches for compositional generalization.\\
\textbf{Compositional Data Augmentation Approaches}: Note that our method also falls into this category. We compare our proposed CompSub and LCS with GECA~\cite{geca} and LexSym~\cite{lexsym} on SCAN, COGS and GeoQuery tasks, Prim2PrimX~\cite{mutual} (in combination with a mutual exclusive training technique) on SCAN and COGS tasks, Comp-IBT~\cite{compibt} on the SCAN tasks, and SUBS~\cite{subs} on the GeoQuery tasks.\\
\textbf{Methods with Specifically Designed Architectures}: We include multiple baseline methods that design specific architectures for compositional generalization tasks, such as CGPS~\cite{primsub}, LexLearn~\cite{lexlearn}, IR-Transformer~\cite{IR-transf}, Dangle~\cite{dangle} and SpanParse~\cite{spanparse}.\\
\textbf{Methods That Leverage Special Training Algorithms}: We also consider using meta-learning training method MAML~\cite{meta-comp} and mutual exclusive training technique~\cite{mutual} to improve compositional generalization as two baselines.\\
\textbf{Pre-trained Language Models}: Apart from the aforementioned approaches which are specialized for compositional generalization, we also include several widely-used pre-trained language models including T5~\cite{t5}, BART~\cite{bart} and GPT-3.5~\cite{ouyang2022traininglanguagemodelsfollow} (code-davinci-002) in the experiment results for readers' reference.\\
\textbf{Prompting Techniques}: For ICL experiments, we adopt different types of prompting techniques as our baselines. For \textbf{\textit{basic prompting techniques}}, we select standard few-shot ICL~\cite{gpt3} and few-shot Chain-of-Thought (CoT)~\cite{chain_of_thought} prompt as our baselines. We also select two approaches, Primitive Coverage~\cite{an-etal-2023-context} and BM25~\cite{bm25} that are selecting few-shot demonstrations that are most \textbf{\textit{relevant}} to the testing case. On their basis, we adopt three approaches, Cosine-Score Retrieval (CSR)~\cite{gupta-etal-2023-coverage}, Bert-Score Retrieval (BSR)~\cite{gupta-etal-2023-coverage} and Gist-Score Retrieval (GSR)~\cite{gupta2024gistscore}, that are selecting not only relevant but also most \textbf{\textit{informative}} demonstrations as the ICL demonstrations. Last but not least, we also extend other state-of-the-art \textbf{\textit{data augmentation}} methods, GECA~\cite{geca} and LexSym~\cite{lexsym}, to the ICL scenarios as our baselines for a comprehensive comparison. For each approach, we test its performance with four different shot numbers: 6, 12, 18 and 24.

Besides, since our proposed methods is \textbf{\textit{model-agnostic}}, in our experiments we adopt diverse base model architectures (note that the comparison between our methods and other baseline methods is with the same base models to guarantee fairness):
(1) Long Short Term Memory (LSTM) -based~\cite{LSTM} + Attention-based Seq-to-Seq~\cite{seq2seq} Models, which is a traditional and widely-used architecture to model sequential data (e.g., text data). We conduct experiments with both one-layer LSTM models aligned with ~\cite{geca} and two-layer LSTM models aligned with ~\cite{lexlearn}.
(2) Vanilla Transformers~\cite{transformer}, which is another powerful architecture to model language. We conduct experiments with three-layer Transformers following the setting in ~\cite{mutual}.
(3) Pre-trained LMs, which are typically pre-trained on large-scale unlabeled text data in a self-supervised manner. We aim to investigate whether our methods consistently boost pre-trained language models' performance or not. We adopt BART-base~\cite{bart} in experiments on the GeoQuery dataset.
(4) LLMs, which are used in our experiments regarding to ICL scenarios. We choose two LLMs that are widely-used in academic works and open-source community: LLaMA-2-13B~\cite{touvron2023llama2openfoundation} and LLaMA-3-8B~\cite{dubey2024llama3herdmodels} as our base models.
\subsection{Main Results}
\label{sec:exp_results}
The main results of our experiments on SCAN, COGS, GeoQuery and COGS-QL tasks are shown in Table~\ref{tab:scan_exps}, Table~\ref{tab:cogs_exps}, Table~\ref{tab:geoquery_exps} and Table~\ref{tab:cofe_icl_prompt} respectively, where with SCAN, COGS and GeoQuery tasks we mainly test the performance of training (or fine-tuning) version of our proposed approaches and with COGS-QL task we mainly test the performance of ICL version of our proposed approaches. 
It is noteworthy that in Table~\ref{tab:scan_exps}, ~\ref{tab:cogs_exps} and ~\ref{tab:geoquery_exps}, “\textbf{+CompSub}" refers to directly leveraging Algorithm~\ref{algo:spansub} to generate additional training data and train (or fine-tune in the case of pre-trained LMs) the model on the original training data and the generated additional data as well; “\textbf{+CompSub+LCS}" refers to leveraging Algorithm~\ref{algo:l2s2} to train models with the LCS framework on the original training data and the additional data generated by CompSub. 
In Table~\ref{tab:cofe_icl_prompt}, “\textbf{+LCS}" refers to utilize the LCS-ICL Algorithm in the LLM and ICL scenarios. 
On SCAN and COGS tasks, we run each experiment with five different seeds and report both of the mean and the standard deviation. On GeoQuery tasks, we run each experiment with three different seeds and report the mean of testing performances. Detailed experiment results on each dataset are listed as follows.\\

\begin{table*}[t]
\caption{
Test accuracy on SCAN Jump, Around Right, Length and MCD splits.
}
\centering
\resizebox{\textwidth}{!}{
\begin{tabular}{lccc|ccc}
\hline
\toprule
\multirow{2}{*}{\textbf{Method}} & \multicolumn{3}{c}{\textbf{SCAN-Original}\footnotesize{~\cite{scan}}} & \multicolumn{3}{c}{\textbf{SCAN-MCD}\footnotesize{~\cite{MCD}}} \\
 & Jump & Around Right & Length & MCD1 & MCD2 & MCD3 \\
\midrule\midrule
CGPS\footnotesize{~\cite{primsub}} &$98.8\%$\tiny{$\pm$ $1.4\%$} & $83.2\%$\tiny{$\pm$ $13.2\%$}& $20.3\%$\tiny{$\pm$ $1.1\%$} &$1.2\%$\tiny{$\pm$ $1.0\%$} &$1.7\%$\tiny{$\pm$ $2.0\%$} &$0.6\%$\tiny{$\pm$ $0.3\%$}   \\
GECA+MAML\footnotesize{~\cite{meta-comp}}  & -- & -- & -- &$58.9\%$\tiny{$\pm$ $6.4\%$}& $34.5\%$\tiny{$\pm$ $2.5\%$}& $12.3\%$\tiny{$\pm$ $4.9\%$} \\
Comp-IBT\footnotesize{~\cite{compibt}}  & $99.6\%$ & $37.8\%$ &$77.7\%$ & $64.3\%$& $80.8\%$& $52.2\% $\\
T5-11B\footnotesize{~\cite{t5}}  & $98.3\%$ & $49.2\% $&$2.0\%$ &$7.9\%$& $2.4\%$&$ 16.2\% $\\
 \midrule 
\textbf{Seq-to-Seq}\footnotesize{~\cite{seq2seq}}  &$1.3\%$\tiny{$\pm$ $0.4\%$}& $10.2\%$\tiny{$\pm$ $4.6\%$} &$12.5\%$\tiny{$\pm$ $2.5\%$} &$8.9\%$\tiny{$\pm$ $1.6\%$} &$11.9\%$\tiny{$\pm$ $9.4\%$}  & $6.0\%$\tiny{$\pm$ $0.9\%$}  \\
+GECA\footnotesize{~\cite{geca}} &$95.2\%$\tiny{$\pm$ $8.0\%$} & $84.3\%$\tiny{$\pm$ $6.3\%$}&$12.8\%$\tiny{$\pm$$ 2.1\%$} &$23.4\%$\tiny{$\pm$ $9.1\%$} &$25.5\%$\tiny{$\pm$ $8.8\%$}  & $10.9\%$\tiny{$\pm$ $4.6\%$}  \\
+LexLearn\footnotesize{~\cite{lexlearn}} &$91.2\%$\tiny{$\pm$ $11.9\%$} & $95.3\%$\tiny{$\pm$$ 1.6\%$} & $13.8\%$\tiny{$\pm$$ 3.0\%$} &$12.5\%$\tiny{$\pm$ $2.0\%$} &$ 19.3\%$\tiny{$\pm$ $1.9\%$} & $ 11.6\%$\tiny{$\pm$ $0.9\%$}  \\
+LexSym\footnotesize{~\cite{lexsym}} &$100.0\%$\tiny{$\pm$ $0.0\%$} &$84.0\%$\tiny{$\pm$$ 7.1\%$} &$14.3\%$\tiny{$\pm$$ 2.7\%$} &$47.4\%$\tiny{$\pm$ $7.1\%$} &$30.8\%$\tiny{$\pm$ $8.4\%$}  & $13.7\%$\tiny{$\pm$ $3.6\%$}  \\
+Prim2PrimX+MET\footnotesize{~\cite{mutual}}  &$7.3\%$\tiny{$\pm$ $5.6\%$} & $97.6\%$\tiny{$\pm$ $1.0\%$}&$15.2\%$\tiny{$\pm$$ 4.5\%$} &$ 31.5\%$\tiny{$\pm$ $4.1\%$}&$ 33.5\%$\tiny{$\pm$ $2.7\%$}&  $11.6\%$\tiny{$\pm$ $1.0\%$} \\
+GECA+MAML\footnotesize{~\cite{meta-comp}}  & $\underline{95.8}\%$\tiny{$\pm$ $6.9\%$}& $86.2\%$\tiny{$\pm$ $5.6\%$}&$13.5\%$\tiny{$\pm$$ 2.2\%$} & $28.2\%$\tiny{$\pm$ $9.6\%$}& $31.8\%$\tiny{$\pm$ $8.5\%$}& $11.2\%$\tiny{$\pm$ $4.2\%$} \\
+CompSub \footnotesize{(\textbf{Ours})} &{$\bm{100.0\%}$}\tiny{$\pm$ $0.0\%$} &$\underline{99.9}\%$\tiny{$\pm$$ 0.1\%$} &$\underline{85.6}\%$\tiny{$\pm$ $8.8\%$} &$\underline{63.4}\%$\tiny{$\pm$ $13.1\%$} &$\underline{72.9}\%$\tiny{$\pm$ $10.1\%$}  & $\underline{74.0}\%$\tiny{$\pm$ $10.2\%$}  \\
+CompSub+LCS \footnotesize{(\textbf{Ours})}  &$\bm{100.0\%}$\tiny{$\pm$ $0.0\%$} &$\bm{100.0\%}$\tiny{$\pm$ $0.0\%$} &$\bm{97.1\%}$\tiny{$\pm$ $3.2\%$} &$\bm{67.4\%}$\tiny{$\pm$ $12.1\%$} &$\bm{73.0\%}$\tiny{$\pm$ $10.1\%$}  & $\bm{80.2\%}$\tiny{$\pm$ $1.8\%$}  \\
 \midrule 
\textbf{Transformer}\footnotesize{~\cite{transformer}}  &$3.5\%$\tiny{$\pm$ $1.7\%$}& $19.9\%$\tiny{$\pm$ $10.4\%$} &$12.2\%$\tiny{$\pm$ $5.6\%$} &$1.7\%$\tiny{$\pm$ $0.7\%$} &$4.3\%$\tiny{$\pm$ $1.3\%$}  & $4.4\%$\tiny{$\pm$ $1.2\%$}  \\
+GECA\footnotesize{~\cite{geca}} &$90.8\%$\tiny{$\pm$ $2.6\%$}& $88.8\%$\tiny{$\pm$ $2.0\%$}& $13.5\%$\tiny{$\pm$ $6.3\%$} &$5.2\%$\tiny{$\pm$ $1.4\%$} &$8.4\%$\tiny{$\pm$ $1.7\%$}  & $7.8\%$\tiny{$\pm$ $2.0\%$}  \\
+LexSym\footnotesize{~\cite{lexsym}}&$91.7\%$\tiny{$\pm$ $0.7\%$}& $90.5\%$\tiny{$\pm$ $3.1\%$} & $16.7\%$\tiny{$\pm$ $3.4\%$} &$19.0\%$\tiny{$\pm$ $2.8\%$} &$78.9\%$\tiny{$\pm$ $6.4\%$}  & $61.0\%$\tiny{$\pm$ $5.5\%$}  \\
+Prim2PrimX+MET\footnotesize{~\cite{mutual}}  &$62.5\%$\tiny{$\pm$ $15.5\%$}& $62.4\%$\tiny{$\pm$ $24.4\%$}& $26.2\%$\tiny{$\pm$ $8.3\%$} &$20.5\%$\tiny{$\pm$ $3.2\%$} &$77.4\%$\tiny{$\pm$ $10.7\%$}  & $58.5\%$\tiny{$\pm$ $2.7\%$}  \\
+CompSub\footnotesize{(\textbf{Ours})}&$\bm{92.4}\%$\tiny{$\pm$ $1.1\%$}& $\underline{92.1}\%$\tiny{$\pm$ $1.5\%$} &$\underline{93.5}\%$\tiny{$\pm$ $0.9\%$} &$\underline{24.8}\%$\tiny{$\pm$ 1.7\%} & $\underline{79.4}\%$\tiny{$\pm$ 1.5\%}  & $\underline{61.3}\%$\tiny{$\pm$ 0.9\%}\\
+CompSub+LCS \footnotesize{(\textbf{Ours})}  &$\underline{92.2}\%$\tiny{$\pm$ $1.6\%$}& $\bm{93.9}\%$\tiny{$\pm$ $1.5\%$} &$\mathbf{95.4\%}$\tiny{$\pm$ $1.0\%$} &$\mathbf{27.0\%}$\tiny{$\pm$ $4.4\%$} & $\mathbf{80.2\%}$\tiny{$\pm$ 1.9\%}  & $\mathbf{63.3\%}$\tiny{$\pm$ 2.3\%}\\
\bottomrule
\end{tabular}
}
\label{tab:scan_exps}
\end{table*}

\textbf{SCAN Results} In Table~\ref{tab:scan_exps}, we observe that our proposed method CompSub leads to significant improvement (in comparison with existing baselines) for both of Seq-to-Seq and Transformer architectures on almost all of six sub-tasks that we study (\textit{jump}, \textit{around right}, \textit{length}, \textit{MCD1}, \textit{MCD2} and \textit{MCD3}). 
We find that on the easier testing tasks \textit{jump} and \textit{around right} CompSub can reach near perfect prediction accuracy (near $100.0\%$), while on the more difficult tasks~\cite{MCD} \textit{length}, \textit{MCD1}, \textit{MCD2} and \textit{MCD3} though achieving tremendous improvement over the base model (by at most $73.1\%$) there still exist large room for further improvement. We observe that exactly on these harder tasks, with the LCS training framework we can consistently further achieve generalization performance gain on the basis of solely adopting CompSub, which also implies that LCS can help model focus more on these difficult compositional patterns. Specifically, on the \textit{length} and \textit{MCD3} sub-tasks, adopting LCS can additionally bring improvement of 11.5\% and 6.2\% on average. 
Comparing our method and exisiting methods, we observe that the combination of CompSub and LCS exceeds most baseline approaches by a large margin. The only exception is the performance on the \textit{MCD2} tasks with Comp-IBT is a little bit higher than the performance with Seq-to-Seq architecture and CompSub. Note that Comp-IBT requires to access $30\%$ compositional generalization testing data in advance (in the training phase) and hence it is not directly comparable with ours.\\

\begin{table}[t]
\caption{
Overall test accuracy on COGS dataset.
}
\centering
\resizebox{0.40\textwidth}{!}{
\begin{tabular}{lc}
\toprule
Method & COGS\scriptsize{~\cite{cogs}}\\
\midrule \midrule
MAML\scriptsize{~\cite{meta-comp}}  &$  64.1\% \tiny{\pm 3.2\%}  $\\
IR-Transformer\scriptsize{~\cite{IR-transf}}   & $  78.4\% $\\
Roberta+Dangle\scriptsize{~\cite{dangle}}   & $   87.6\%$ \\
T5-Base\scriptsize{~\cite{t5}}   &  $  85.9\% $\\
\midrule
\textbf{Seq-to-Seq}\scriptsize{~\cite{seq2seq}}  & $ 55.4\% \tiny{\pm 4.2\%} $\\
\midrule
+GECA\scriptsize{~\cite{geca}}   &  $ 48.0\% $\tiny{$\pm 5.0\%$} \\
+LexLearn\scriptsize{~\cite{lexlearn}}  & $ 82.0\%$ \tiny{$\pm 0.0\%$} \\
+LexSym\scriptsize{~\cite{lexsym}}  &  $ 81.4\% $\tiny{$\pm 0.5\%$}  \\
+Prim2PrimX+MET\scriptsize{~\cite{mutual}}  &$ 81.1\% $\tiny{$\pm 1.0\%$} \\
+CompSub \footnotesize{(\textbf{Ours})}  & $\underline{91.8\%} $\tiny{$\pm 0.1\%$} \\
+CompSub+LCS \footnotesize{(\textbf{Ours})} & $\bm{92.3\%}$\tiny{$\pm 0.2\%$}  \\
\bottomrule
\end{tabular}
}
\label{tab:cogs_exps}
\end{table}

\textbf{COGS Results} In Table~\ref{tab:cogs_exps}, on the COGS task the performance of our base model (Seq-to-Seq) increases from $55.4\%$ to $91.8\%$ when we leverage CompSub to generate additional training data. CompSub has approximately $10\%$ lead compared with our baseline methods (LexLearn, LexSym, Prim2PrimX+MET) implemented on the exact same base model. Even compared with methods that leverage powerful pre-trained language models (e.g., Roberta+Dangle and T5-Base), LSTM+CompSub still has some advantages.
Furthermore, we can improve the performance of our base model from 91.8$\%$ to 92.3$\%$ when adopting the LCS training framework based on CompSub.\\

\begin{table}[h!]
\caption{
Test accuracy on GeoQuery question (i.i.d.) and query (compositional) splits.
}
\centering
\resizebox{0.48\textwidth}{!}{
\begin{tabular}{lc|c}
\toprule
\multirow{2}{*}{\textbf{Method}} & \multicolumn{2}{c}{\textbf{GeoQuery}\footnotesize{~\cite{geoquery,template-split}}}\\
 & Question & Query \\
\midrule \midrule
SpanParse\scriptsize{~\cite{spanparse}}& $78.9\%$ & $76.3\%$\\
code-davinci-002\scriptsize{~\cite{ouyang2022traininglanguagemodelsfollow}}+Cover-LS\scriptsize{~\cite{levy-etal-2023-diverse}}& $88.7\%$ & $85.3\%$\\
\midrule 
\textbf{Seq-to-Seq}\scriptsize{~\cite{seq2seq}} &$ 75.2\%$ & $58.6\%$\\
+GECA\scriptsize{~\cite{geca}} & $76.8\%$ &  $60.6\%   $\\
+LexSym\scriptsize{~\cite{lexsym}} &$81.6\%$ & $80.2\% $   \\
+SUBS\scriptsize{~\cite{subs}}  &$80.5\%$ &  $77.7\%  $ \\
+CompSub \footnotesize{(\textbf{Ours})}&\underline{$82.4\%$} & \underline{$81.4\%$}\\
+CompSub+LCS \footnotesize{(\textbf{Ours})}&$ \bm{83.2\%}$ & $\bm{82.2\%} $\\
\midrule
\textbf{BART}\scriptsize{~\cite{bart}} &$90.2\%$ & $71.9\%$\\
+GECA\scriptsize{~\cite{geca}} & $87.9\%$ &   $83.0\%  $\\
+LexSym\scriptsize{~\cite{lexsym}} &$90.2\%$ & $87.7\%  $  \\
+SUBS\scriptsize{~\cite{subs}} & $\bm{91.8\%}$ &  $88.3\%   $\\
+CompSub \footnotesize{(\textbf{Ours})} & \underline{$90.6\%$} & \underline{$89.5\%$} \\
+CompSub+LCS \footnotesize{(\textbf{Ours})} & \underline{$90.6\%$} & $\bm{89.7\%} $\\
\bottomrule
\end{tabular}
}
\label{tab:geoquery_exps}
\end{table}

\textbf{GeoQuery Results} In Table~\ref{tab:geoquery_exps}, we show that on the compositional template query split CompSub leads to substantial and consistent improvement over other baseline data augmentation methods (GECA, LexSym and SUBS) on both of implementations based on the Seq-to-Seq architecture and the pre-trained BART model, achieving new state-of-the-art results (pushing forward the previously state-of-the-art results by 1.2$\%$). As for the i.i.d question split, CompSub still has advantages over baseline methods when based on the Seq-to-Seq model. When we adopt BART as our base model, CompSub boosts the performance of BART by $0.4\%$ which is ahead of GECA and LexSym, falling behind SUBS.\\

\begin{table*}[t]
\caption{
Test accuracy on COGS-QL with In-Context Learning (Out-of-Distribution Performance).
}
\centering
\resizebox{0.9\textwidth}{!}{
\begin{tabular}{lcccc|cccc}
\hline
\toprule
\textbf{Model} & \multicolumn{4}{c}{\textbf{LLaMA2-13B}} & \multicolumn{4}{c}{\textbf{LLaMA3-8B}} \\
\textbf{Method and Shot} & 6 & 12 & 18 & 24 & 6 & 12 & 18 & 24  \\
\midrule\midrule
\multicolumn{5}{l}{\textit{Basic Prompting Techniques}}\\
+Standard ICL\footnotesize{~\cite{gpt3}}  &$15.0\%$& $17.5\%$  &$24.5\%$ &$30.5\%$ &$18.0\%$& $23.6\%$  &$28.6\%$ &$31.3\%$   \\
+CoT Prompt\footnotesize{~\cite{chain_of_thought}}  &$19.0\%$& $22.0\%$  &$18.0\%$ &$21.5\%$ &$21.3\%$& $25.8\%$  &$28.9\%$ &$29.6\%$   \\
\midrule
\multicolumn{5}{l}{\textit{Selecting Relevant Demonstrations}}\\
+PrimCoverage\footnotesize{~\cite{an-etal-2023-context}}  &$34.5\%$& $39.5\%$  &\underline{$43.0\%$} &$44.5\%$  &$43.5\%$& $48.3\%$  &$48.8\%$ &$50.0\%$    \\
+BM25\footnotesize{~\cite{bm25}} &$12.5\%$& $18.0\%$  &$21.0\%$ &$17.0\%$  &$19.7\%$& $24.9\%$  &$33.3\%$ &$28.6\%$   \\
\midrule
\multicolumn{5}{l}{\textit{Selecting Informative Demonstrations}}\\
+CSR\footnotesize{~\cite{gupta-etal-2023-coverage}} &$39.5\%$& $41.5\%$  &$44.0\%$ &$44.0\%$ &$27.8\%$& $32.4\%$  &$31.7\%$ &$32.8\%$   \\
+BSR\footnotesize{~\cite{gupta-etal-2023-coverage}} &$\underline{42.0}\%$& $\underline{43.0}\%$  &$\underline{44.0}\%$ &$44.5\%$  &$33.0\%$& $35.1\%$  &$37.0\%$ &$37.8\%$  \\
+GSR\footnotesize{~\cite{gupta2024gistscore}} &$\bm{42.5}\%$& $\bm{44.0}\%$  &$\underline{44.0}\%$ &$42.5\%$  &$29.8\%$& $33.9\%$  &$35.1\%$ &$36.0\%$   \\
\midrule
\multicolumn{5}{l}{\textit{Augmenting Demonstrations}}\\
+GECA\footnotesize{~\cite{geca}}  &$30.5\%$& $35.0\%$  &$37.0\%$ &$38.5\%$ &$40.6\%$& $44.8\%$  &$45.6\%$ &$46.3\%$   \\
+LexSym\footnotesize{~\cite{lexsym}}  &$29.0\%$& $36.0\%$  &$39.5\%$ &$40.5\%$   &$41.7\%$& $46.1\%$  &$47.5\%$ &$47.7\%$  \\
+LCS-ICL(\textit{random version}) &$35.5\%$& $39.5\%$  &$42.0\%$ &\underline{$46.0\%$}  &\underline{$45.3\%$}& \underline{$50.5\%$}  & \underline{$54.4\%$} & \underline{$55.9\%$}  \\
+LCS-ICL(\textit{learned version}) &$37.0\%$ & $42.0\%$  &$\mathbf{44.5}\%$ &$\mathbf{47.5}\%$ &$\mathbf{45.5\%}$& $\mathbf{52.2}\%$  &$\mathbf{57.6}\%$ &$\mathbf{58.7\%}$   \\
\bottomrule
\end{tabular}
}
\label{tab:cofe_icl_prompt}
\end{table*}

\textbf{COGS-QL Results} In Table~\ref{tab:cofe_icl_prompt}, we show the empirical results of utilizing our LCS algorithm to boost \textit{LLMs' in-context learning performance} on the COGS-QL task. Firstly, we find that directly using standard few-shot ICL prompting or chain-of-thought ICL prompting can hardly solve the COGS-QL task. We clearly observe that for both of two LLMs (LLaMA2-13B and LLaMA3-8B) using more in-context demonstrations in our LCS approach leads to better performance. However, many existing baselines (e.g., CSR, GSR, LexSym and stuff) do not show such an monotone increasing trend. When comparing our approach (LCS) with existing baselines, we find that in most cases \textbf{\textit{LCS outperform best baseline methods}} by at most $8.8\%$ using LLaMA3-8B and 18-shot ICL prompt: best baseline PrimCoverage ($48.8\%$) versus our LCS ($57.6\%$). Meanwhile we also note that with LLaMA2-13B and small shot numbers (6 and 12) the performance of LCS is a little bit worse than best baselines (GSR) while when shot number increases (18 and 24) LCS surpasses GSR (e.g., shot number $=24$: LCS ($47.5\%$)) versus GSR ($42.5\%$). This is mainly because the small number of ICL demonstrations limits the initial search space of LCS, hence weakening its performance.

To draw a short conclusion, we conduct extensive experiments and empirically demonstrate the superiority of CompSub and LCS from the perspective of overall compositional generalization performance. In this process, we consider different tasks, different base model architectures and different shot numbers of in the ICL prompts to ensure the rigor and the comprehensiveness of our study.
\subsection{Result Interpretation, Analysis and Ablation Study}
\label{sec:exp_analysis}
In this section, we mainly present results for some analysis experiments. Specifically, we aim to further answer the following Research Questions (RQs).
\textbf{RQ1}: (\textit{Analyzing the performance of CompSub}) Grounded on the same original training set, does the CompSub help with fully exploring the potential augmentation space as supposed in Section~\ref{sec:intro}, especially when compared with existing augmentation approaches~\cite{geca,subs,lexsym}?
\textbf{RQ2}: (\textit{Analyzing the performance of LCS}) Is the LCS truly capable of automatically mining hard and rare data re-combination patterns as supposed in Section~\ref{sec:intro}? What kind of re-combination patterns are recognized as difficult for language models?
\textbf{RQ3}: (\textit{Ablation Studies}) One of the key modules in the LCS framework (both for the Algorithm~\ref{algo:l2s2} and the LCS-ICL Algorithm in our design is the selective augmentation module. Does this up-stream selective module really play a necessary role in compositional generalization?
\textbf{RQ4}: (\textit{Applicability to different base models}) Are our proposed methods applicable to different architectures (e.g., for sequence modeling architecture: LSTM~\cite{LSTM} and Transformer~\cite{transformer}; for pre-trained language model: BART~\cite{bart}, LLaMA2~\cite{touvron2023llama2openfoundation} and LLaMA3~\cite{dubey2024llama3herdmodels})?\\
\textbf{RQ1: Analyzing the performance of CompSub} 
To help deepen the understanding of the performance improvement brought by CompSub, we first break down the overall performance on COGS task into four different parts (for specific examples please refer to Table~\ref{tab:cogs_gen_examples}), including lexical generalization performance and three different types of structural generalization performance. 
\begin{table*}[t]
\caption{
Specific examples of different generalization types in COGS. We include a lexical generalization case and three structural generalization cases (corresponding to obj\_pp\_to\_sub\_pp, pp\_recursion and cp\_recursion, respectively).
}
\centering
\resizebox{0.95\textwidth}{!}{
\begin{tabular}{lcc}
\toprule
\textbf{generalization type} & \textbf{training} & \textbf{generalization} \\
\midrule
lexical & \textbf{Lina} gave the cake to Olivia.  & The cat liked \textbf{Lina}. \\
\midrule
obj\_pp\_to\_sub\_pp & Noah ate \textbf{the cake on the plate}. & \textbf{The cake on the table} burned.\\
\midrule
pp\_recursion & Ava saw the ball \textbf{in the bottle on the table}. & Ava saw the ball \textbf{in the bottle on the table on the floor}.\\
\midrule
cp\_recursion & Emma said \textbf{that} Noah knew \textbf{that} the cat danced. & Emma said \textbf{that} Noah knew \textbf{that} Lucas saw \textbf{that} the cat danced.\\
\bottomrule
\end{tabular}
}
\label{tab:cogs_gen_examples}
\end{table*}

\begin{table}
\caption{
The maximum numbers of distinct augmented examples on the query split of GeoQuery dataset with different augmentation methods. w/o Aug refers to the number of original training examples.
}
\centering
\resizebox{0.47\textwidth}{!}{
\begin{tabular}{c|c|c|c|c}
\toprule
\footnotesize{w/o Aug} & \footnotesize{GECA} &  \footnotesize{LexSym} & \footnotesize{SUBS}& \footnotesize{CompSub} \\
\midrule
$519$ & $2,028$ & $28,520$ & $20,564$ & $99,604$ \\
\bottomrule
\end{tabular}
}
\label{tab:max_aug_num}
\end{table}

\begin{table}
\caption{
Test accuracy of different generalization types in COGS task. "lex" refers to lexical generalization test; "s1","s2" and "s3" refer to "obj\_pp\_to\_subj\_pp", "pp\_recursion", "cp\_recursion" respectively, which are 3 different types of structural generalization tests.
}
\centering
\resizebox{0.48\textwidth}{!}{
\begin{tabular}{lc|c|c|c}
\toprule
\multirow{2}{*}{\textbf{Method}} & \multicolumn{4}{c}{\textbf{Testing Type in COGS}\footnotesize{~\cite{cogs}}}\\
& lex & s1 & s2 & s3 \\
\midrule
\footnotesize{Seq-to-Seq}\footnotesize{~\cite{seq2seq}} & \footnotesize{$69.3\%$} & \footnotesize{$0.0\%$} & \footnotesize{$0.0\%$} & \footnotesize{$0.9\%$}\\
\footnotesize{+LexSym} & \footnotesize{$95.3\%$} & \footnotesize{$0.0\%$} & \footnotesize{$0.0\%$} & \footnotesize{$0.7\%$}\\ 
\footnotesize{+CompSub} & \footnotesize{$99.1\%$} & \footnotesize{$91.8\%$} & \footnotesize{$45.0\%$} & \footnotesize{$7.2\%$}\\
\footnotesize{+CompSub+LCS} & \footnotesize{$99.4\%$} & \footnotesize{$93.7\%$} & \footnotesize{$45.1\%$}& \footnotesize{$10.7\%$}\\
\bottomrule
\end{tabular}
}
\label{tab:cogs_type}
\end{table}

\begin{figure}
\centering 
\includegraphics[width=0.47\textwidth]{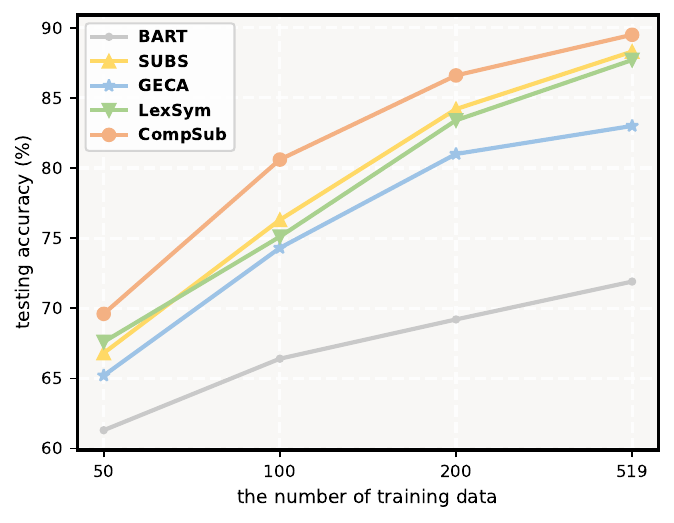}
\caption{Performance of CompSub in the low-data regime, in comparison with existing augmentation baselines: GECA, LexSym and SUBS.}
\label{figure:fewshot}
\vspace{-0.1in}
\end{figure}

The results are shown in Table~\ref{tab:cogs_type}.
Compared with LexSym, which only enable single-grained substitutions (i.e., substituting for single words), we find that CompSub can not only improve
generalization on testing cases of different structural types, but also further boost the lexical level generalization. 
We have similar observation in Table~\ref{tab:scan_exps}. 
In SCAN results, we find that for lexical generalization (i.e., \textit{jump split}), CompSub achieves similar performance with the best augmentation baseline LexSym. 
When it comes to diverse and more difficult structural generalization cases (e.g., \textit{MCD1}, \textit{MCD2} and \textit{MCD3}), CompSub exceeds existing augmentation baselines (GECA, LexSym and Prim2PrimX) by a very large margin.
Besides, in Section~\ref{sec:intro}, we hypothesize that CompSub enables multi-grained compositions of substantial substructures in the whole training set
and thus lead to improvement for various kinds of compositional generalization. 
We provide a statistic on the maximum number of augmented examples (after deduplication) on the query split of GeoQuery dataset (the original training set only contains $519$ examples.) with different augmentation methods, including GECA, LexSym, SUBS and CompSub in Table~\ref{tab:max_aug_num}. 
It is noteworthy that CompSub (augmenting to near $100,000$ examples) overwhelmingly outweighs other augmentation methods (less than $30,000$ examples), which reflects its superiority of exploring potential compositions of substantial substructures in the entire training set.
We further investigate the performance of CompSub in the extremely low-data regime: sampling $50$, $100$ and $200$ examples from the GeoQuery training set, augmenting with CompSub and testing the trained models' performance. We plot the performance (in comparison with existing augmentation baselines) in Figure~\ref{figure:fewshot}, demonstrating that our method consistently outperform the baselines in low-data regime.\\
\textbf{RQ2: Analyzing the performance of LCS}
Our main question is \textit{Is the LCS truly capable of automatically mining hard and rare data re-combination patterns?}.

\begin{table}
\caption{
Comparision of the error rates($\downarrow$) of examples with different concepts (i.e., spans) between the random augmentation baseline and LCS (learning augmentation). Results are attained using the same LSTM architecture with ~\cite{geca} on SCAN-MCD3 split.
}
\centering
\resizebox{0.47\textwidth}{!}{
\begin{tabular}{lccc}
\toprule
\multirow{2}{*}{\textbf{Method}} & \multicolumn{3}{c}{\textbf{Error Type in SCAN-MCD~\cite{MCD}}}\\
 & walk right & walk opposite right & walk around right\\
\midrule
Random Aug. & $51.2\%$&$28.1\%$ & $76.8\%$ \\
Learning Aug. & $37.4\%$ & $14.6\%$ & $40.2\%$\\
\bottomrule
\end{tabular}
}
\label{tab:w_a_r_err_dec}
\end{table}

For results on SCAN-MCD tasks, we present the results of leveraging CompSub in Figure~\ref{figure:intro} (d) and observe an imbalanced prediction error rate distribution: the concepts in the SCAN dataset has distinct difficulty.
As a comparison, we present the results of using the LCS augmentation and a random augmentation baseline (which is realized by substituting the learned augmentation module in the LCS framework with a random augmentation module.) in Table~\ref{tab:w_a_r_err_dec}. 
We can conclude that LCS is especially helpful for improving the performance of down-stream neural seq-to-seq models on the prediction of harder examples (here harder examples refer to those contain the compositions of elusive concepts like “walk around right" and novel surroundings).

\begin{figure}
\centering 
\includegraphics[width=0.47\textwidth]{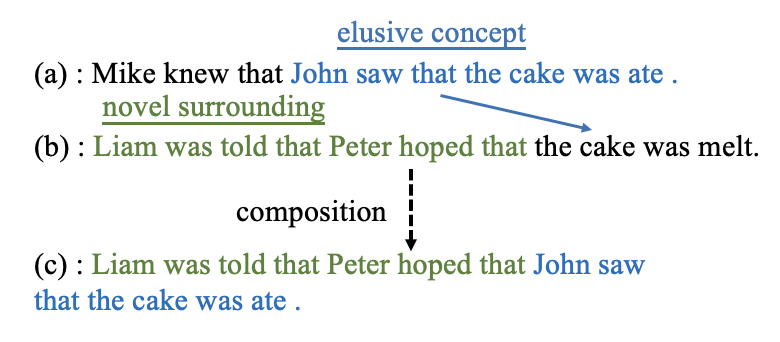}
\caption{A composition that helps to improve “cp\_recursion" generalization. The composition of “Liam was told that Peter hoped that" and “John saw that the cake was ate ." results in deeper recursion of that-structure: the recursion depth increases from 2 to 3.}
\label{figure:cogs_that_case}
\vspace{-0.1in}
\end{figure}

For results on the COGS task: as shown in Table~\ref{tab:cogs_type}, we find that the utilization of LCS framework training can help models better generalize on testing cases of "cp\_recursion" type, which requires precisely combining two special structures (we depict an example in Figure~\ref{figure:cogs_that_case}: "cp\_recursion" requires to precisely combine “John knew that
the cake was ate ." in (a) and “Liam was told that Peter hoped that" in (b) to generate (c) which contains a deeper recursion of \textbf{that}-structure.\\
\textbf{RQ3: The importance of adopting the selective augmentation module in the LCS framework} 
We aim to leverage control experiments to verify the effectiveness of the selective augmentation module in the LCS framework: as for the experimental group, we select the augmented examples \textit{according to the probability output by the learned neural network} in Algorithm~\ref{algo:l2s2} or \textit{directly the feedback output by the down-stream LLMs} in LCS-ICL Algorithm; as for the control group, we select the augmented examples \textit{in a completely random manner}.
For Algorithm~\ref{algo:l2s2}, we show the comparison between the experimental group (denoted by “+Learned Augmentation") and the control group (denoted by “+Random Augmentation") on three SCAN-MCD splits in Table~\ref{tab:different_archs_exps}.

\begin{table}
\caption{
The importance of adopting the selective augmentation module in the LCS framework: comparison between the experimental group (denoted by “+Learned Augmentation") and the control group (denoted by “+Random Augmentation") on three SCAN-MCD splits.
}
\centering
\resizebox{0.47\textwidth}{!}{
\begin{tabular}{lccc}
\toprule
\multirow{2}{*}{\textbf{Method}} & \multicolumn{3}{c}{\textbf{SCAN-MCD}\footnotesize{~\cite{MCD}}}\\
 & MCD1 & MCD2 & MCD3  \\
\midrule
\midrule
\textbf{\emph{Seq-to-Seq}}\footnotesize{~\cite{seq2seq}} &8.9\% &11.9\%  & 6.0\%  \\
\midrule
+Random Augmentation &46.6\% &52.3\%  & 58.8\% \\
+Learned Augmentation & \textbf{55.1\%} & \textbf{54.3\%}\  & \textbf{70.8\%} \\
\midrule
+CompSub &63.4\% &72.9\%  & 74.0\%\\
+CompSub+Random Aug. &63.3\%\ &66.2\%  & 71.2\% \\
+CompSub+Learned Aug. &\textbf{67.4\%} &\textbf{73.0\%}  & \textbf{80.2\%} \\
\midrule
\midrule
\emph{\textbf{Transformer}} \footnotesize{~\cite{transformer}} &1.7\%& 4.3\%  & 4.4\% \\
\midrule
+Random Augmentation &11.2\%& 37.0\% & 48.1\%\\
+Learned Augmentation &\textbf{19.3\%} & \textbf{68.1\%}  & \textbf{57.8\%}\\
\midrule
+CompSub &24.8\% & 79.4\%  & 61.3\%\\
+CompSub+Random Aug. &21.0\% & \textbf{80.2\%}  & 60.3\%\\
+CompSub+Learned Aug. &\textbf{27.0\%} & \textbf{80.2\%}  & \textbf{63.3\%}\\
\bottomrule
\end{tabular}
}
\label{tab:different_archs_exps}
\end{table}

\begin{table}[t]
\caption{
Comparing the performance of two other versions of LCS-ICL algorithm on LLaMA3-8B. The first row represents the results for shuffling the order of the demonstrations that are selected by the original LCS-ICL algorithm. The second row represents the results for the LCS-ICL algorithm that select all of the demonstrations in the fine screening stage at once instead of selecting in a successive manner.
}
\centering
\resizebox{0.48\textwidth}{!}{
\begin{tabular}{l|cccc}
\hline
\toprule
\textbf{Shot} & 6 & 12 & 18 & 24  \\
\midrule
\textbf{\textit{shuffling}} &$46.3\%$\tiny{$\pm{0.4\%}$}  &$52.6\%$\tiny{$\pm{0.7\%}$}  &$57.2\%$\tiny{$\pm{0.7\%}$} &$58.6\%$\tiny{$\pm{0.4\%}$} \\
\textbf{\textit{w.o. succ. select.}} &$43.8\%$& $52.2\%$  &$54.8\%$ &$56.9\%$  \\
\bottomrule
\end{tabular}
}
\label{tab:cofe_icl_prompt_comparison}
\end{table}

Through observing the results, we conclude that (for both Seq-to-Seq~\cite{seq2seq} and Transformer~\cite{transformer} architectures; directly use or use in combination with CompSub) “+Learned Augmentation" groups consistently exhibit better performance than the corresponding “+Random Augmentation" groups, highlighting the importance of adopting the selective augmentation module in Algorithm~\ref{algo:l2s2}.

For the LCS-ICL Algorithm, we show the comparison between the experimental group (denoted by “LCS(\textit{selective version})") and the control group (denoted by “LCS(\textit{random version})") on the COGS-QL dataset in Table~\ref{tab:cofe_icl_prompt}.
We find that in most cases “LCS(\textit{selective version})" perform better than “LCS(\textit{random version})", verifying the effectiveness of the selective augmentation module in the in-context learning scenarios. 
We additionally study the effect of the order of the selected demonstrations and the effect of successively selecting demonstrations in the fine screening stage of the LCS-ICL algorithm. We conduct experiments with LLaMA3-8B and show the results in Table~\ref{tab:cofe_icl_prompt_comparison}. “\textit{shuffling}” refers to that we shuffle the order of the demonstrations that are selected with the LCS-ICL algorithm (with three different random seeds). “\textit{w.o. succ. select.}” refers to that we select all of the rest demonstrations in the fine screening stage of LCS-ICL algorithm at once (instead of selecting demonstrations successively).
We observe that “\textit{w.o. succ. select.}” achieves much lower performance than the original LCS-ICL (Table~\ref{tab:cofe_icl_prompt}), demonstrating that the successively selecting demonstrations is necessary in our design. We also find that perturbing the order of selected demonstrations has little effect on the ICL performance.
\\
\textbf{RQ4: Applicability to different base models}
We demonstrate that our methods can be applicable to different base models: (1) we show the CompSub (Algorithm~\ref{algo:spansub}) and the LCS (Algorithm~\ref{algo:l2s2}) results on SCAN tasks with LSTM-based seq-to-seq models~\cite{seq2seq} and Transformer models~\cite{transformer} in Table~\ref{tab:scan_exps}. (2) we show the CompSub (Algorithm~\ref{algo:spansub}) and the LCS (Algorithm~\ref{algo:l2s2}) results on GeoQuery tasks with LSTM-based seq-to-seq models~\cite{seq2seq} and pre-trained BART models~\cite{bart} in Table~\ref{tab:geoquery_exps}. (3) we show the LCS (the LCS-ICL Algorithm) results on COGS-QL tasks with LLaMA2-13B~\cite{touvron2023llama2openfoundation} and LLaMA3-8B~\cite{dubey2024llama3herdmodels} in Table~\ref{tab:cofe_icl_prompt}.

\section{Conclusion}

In this paper, we focus on improving the compositional generalization performance of neural language models from the perspective of data augmentation. Specifically, we propose a new compositional data augmentation method, CompSub, which is the first to explore span-based compositional data augmentation, thus flexibly supporting to inject multi-grained compositional bias into the training data. 
Building on CompSub, we introduce LCS as a differentiable augmentation framework that first enables difficulty-aware composition and is compatible with various downstream language models. 
We further extend the key ideas of CompSub and LCS to in-context learning (ICL) scenarios of large language models (LLMs), proposing LCS-ICL to select the most appropriate demonstrations to enhance the few-shot compositional generalization capacity of state-of-the-art LLMs. 
Moreover, we provide theoretical insights into the CompSub and LCS algorithms, showing that leveraging CompSub is equivalent to introducing an additional regularization term in the optimization objective. This term encourages the model to learn the invariance of concepts. Additionally, using the LCS framework to train neural language models can reduce the Rademacher complexity, thereby deriving a tighter upper bound on the compositional generalization error.
Finally, our comprehensive experimental results on four datasets (across more than ten tasks) and the corresponding analysis strongly demonstrate the empirical effectiveness of our algorithms.
\section*{Acknowledgement}

 
%

\bibliographystyle{IEEEtran}
\bibliography{reference}

\appendix
\section{Details of CompSub (Algorithm 1)}
\label{appendix:data preprocess}
\subsection{Extraction of span alignments}
For SCAN dataset, since there is no off-the-shelf technique to map sequential data in SCAN dataset to tree-form, we slightly the modify algorithm SimpleAlign from \cite{lexlearn} to extract consecutive span alignments for our experiments on SCAN. We denote the input sequence as $x$, the output sequence as $y$, the span, which is going to be extracted from the input sequence, as $v$ and its counterpart in the output sequence as $w$. Basically, we extract a pair of span alignment $(v,w)$ following the maximally restrictive criterion:
\begin{equation}
\begin{split}
& nec.(v,w)=\forall xy. (w\in y)\rightarrow (v\in x)\\
& suff.(v,w)=\forall xy. (v\in x)\rightarrow (w\in y)\\
& C_1(v,w) = nec.(v,w) \land  suff.(v,w)\\
\end{split}
\label{equ:scan_span}
\end{equation}
Both $v$ and $w$ are supposed to be consecutive fragments in the input sequence and output sequence respectively. \\
We additionally apply appropriate relaxations on the top of criterion(\ref{equ:scan_span}) to enable the extraction of more spans: we tolerate many-to-one mapping and one-to-many mapping to some extent to avoid discarding of "[verb] around [direction]" and "[verb] [direction]"(e.g., both of interpretations of "walk around right" and "walk right" cover "TR W"). 
Besides, we manually set the maximum number of words in $v$ to 3 and the maximum number of words in $w$ to 8. 

\begin{figure}[t]
\centering 
\includegraphics[width=0.47\textwidth]{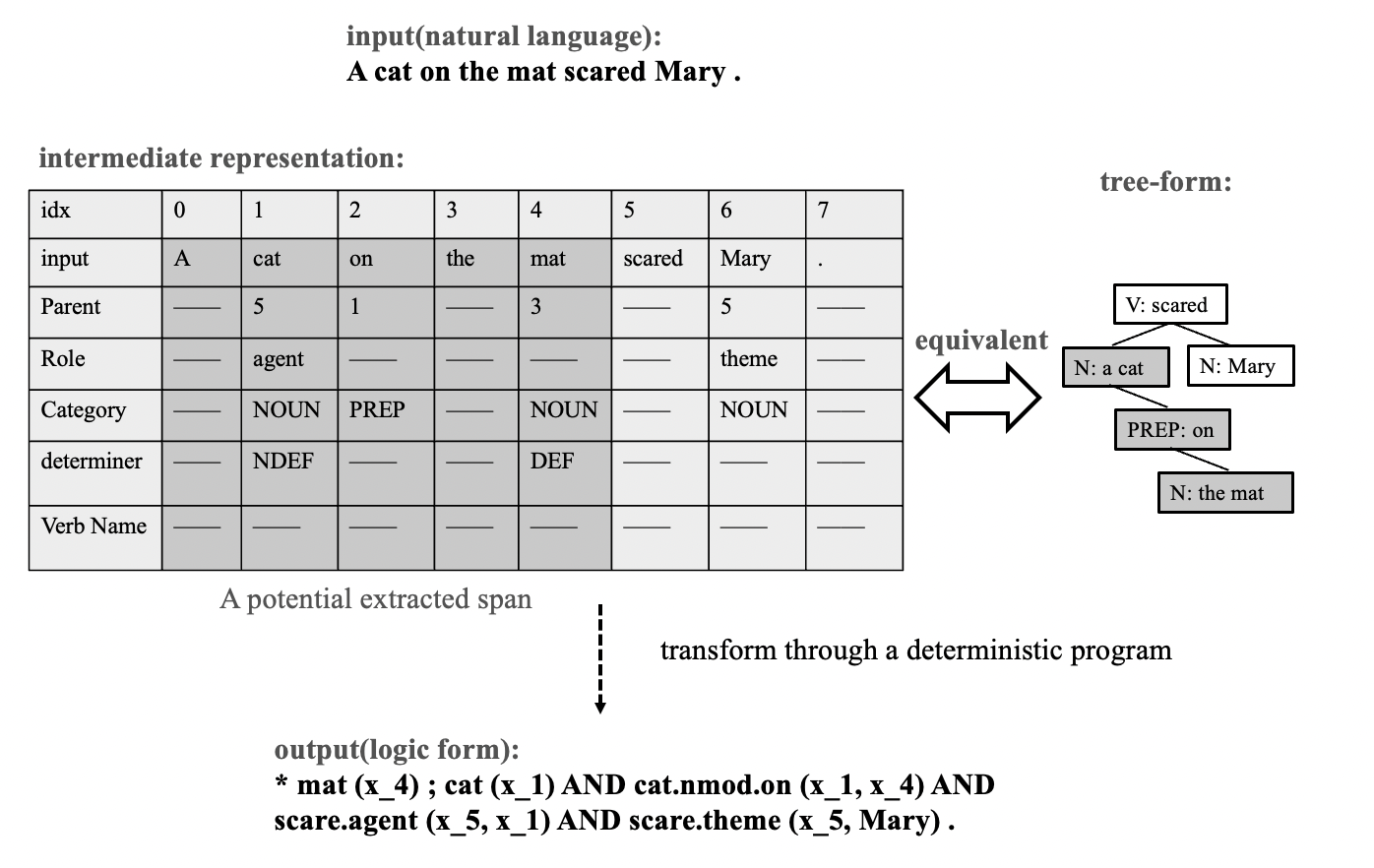}
\caption{An instance for an intermediate representation, its corresponding tree-form and a potential extracted span for COGS.}
\label{figure:cogs_align_instance}
\end{figure}

For COGS, we directly use the intermediate representation from \cite{IR-transf}. An instance of intermediate representation is shown in Fig~\ref{figure:cogs_align_instance}.
We search for every consecutive fragments in the intermediate presentations of COGS to extract eligible spans according to Definition~\ref{define:eligible span}. The naive implementation of the above search algorithm has the time complexity of $\mathcal{O}(n\cdot m^3)$, where $n$ is the number of sentences in the training set and $m$ is the maximal length of a single sentence in the training set.

For GeoQuery, following \cite{subs}, we directly adopt the span trees (\emph{gold trees}) extracted and aligned by ~\cite{spanparse}. And we refer the readers to get more detailed information about how to construct such span trees from the original paper~\cite{spanparse}.\\
Note that we slightly correct several denotations in the original \emph{gold trees} from \cite{spanparse}, for they are slightly differing from the ground-truth. To clarify it, we put an example of modification here (given that the others are similar, we do not present the others here):
\begin{lstlisting}
geoquery["input"] = 
    "what is the population of washington dc ?"
geoquery["program"] = 
    "answer ( population_1 ( cityid ( 'washington', 'dc' ) ) )"
// the original gold_spans
geoquery["gold_spans"] = 
    {"span": [5, 5], "type": "cityid#'washington'"} 
// after correction
geoquery["gold_spans"] = 
    {"span": [5, 6], "type": "cityid#'washington'"} 
    // this is just one of the spans
    // washington dc is the capital city of USA; 
    // washington is a state of USA;
\end{lstlisting}
To ensure a fair comparison with previous substitution-based data augmentation methods~\cite{lexsym,subs}, we rerun their methods on the modified \emph{gold trees}.
\begin{figure}[t]
\centering 
\includegraphics[width=0.47\textwidth]{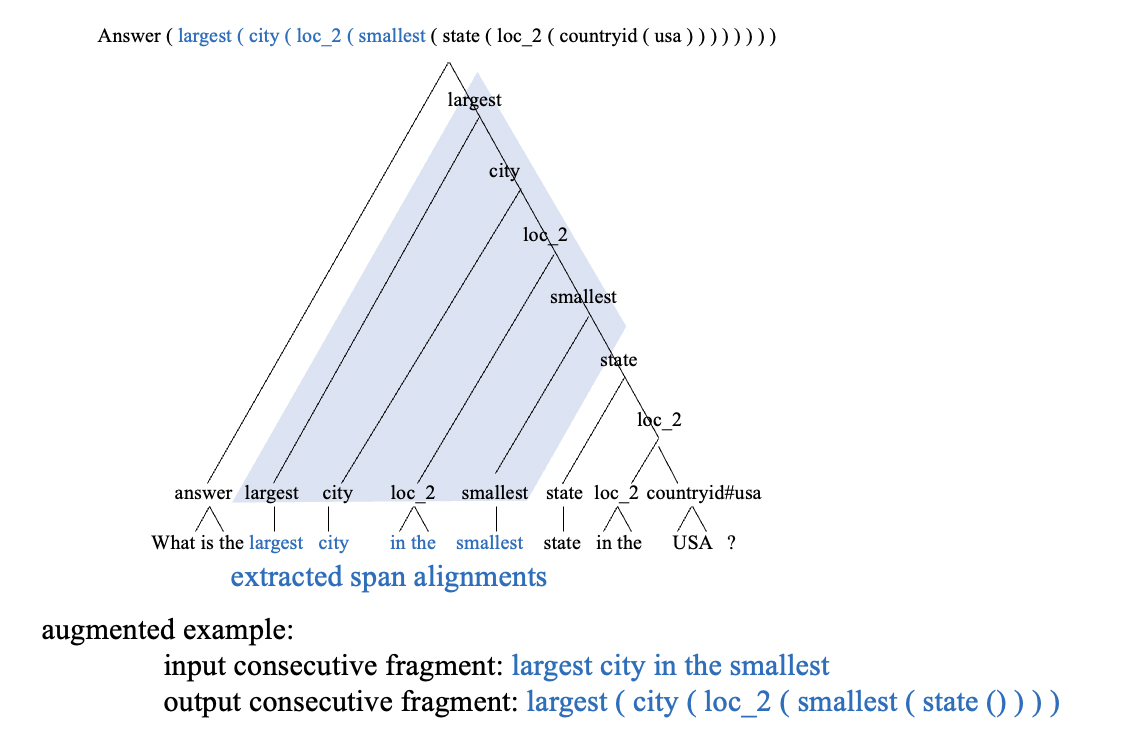}
\caption{An instance for a constructed span tree and extracting a consecutive span from the span tree.}
\label{figure:geoquery_instance}
\end{figure}
\subsection{Inferring the equivalence class of words}
For COGS, we directly leverage the information in the intermediate representations to infer the equivalence class of the words (e.g., NOUN, VERB or PREP). For SCAN and GeoQuery, we use the technique of inferring the types of words form ~\cite{lexsym}, which cluster the words according to their shared contexts in the training set.\\
For GeoQuery, we additionally adopt context2vec methods~\cite{melamud-etal-2016-context2vec} (where we train a simple one-layer LSTM-based mask-reconstruction model) to boost the exploration of potentially syntactically-equivalent words (i.e., candidates to fill in the masked blank). We put the final result of word-clustering on GeoQuery here as follows:(We cluster the words in the target side)\\
\begin{lstlisting}
/*
word clustering result for GeoQuery: 
words not included are not syntactically 
equivalent to any other words
*/
cluster1 = ['highest','major','largest',
            'smallest','shortest','lowest',
            'longest']
cluster2 = ['mountain','state','city',
            'river','place','lake']
cluster3 = ['loc_2','traverse_2']
cluster4 = ['countryid','cityid','stateid',
            'placeid']
cluster5 = ['traverse_1','loc_1','capital_2']
cluster6 = ['largest_one','smallest_one']
cluster7 = ['area_1','density_1','population_1']
cluster8 = ['size','high_point_1']
cluster9 = ['most','fewest']
\end{lstlisting}

\subsection{Substitution Strategy}
In Definition~\ref{define:eligible span}, We present the definition of ``\textit{\textbf{eligible span}}” mentioned in Section \uppercase\expandafter{\romannumeral3}.A.
\begin{myDef}
\label{define:eligible span}
(\textbf{Eligible Span}) Given a sentence or a program sequence ${S}$ = ${[e_0,\ e_1,\ ...,\ e_n]}$, there exists one and only one multi-way tree ${T}$ corresponding to ${S}$, the in-order traversal sequence\footnote{In our case in-order traversal of a multi-way tree is to traverse the most left child, traverse the root node and then traverse left childs from right to left in order.} ${\Lambda}$ of which is ${v_0\rightarrow v_1\rightarrow ...\rightarrow v_n}$ (node ${v_i}$ corresponds to token ${e_i}$, $0 \leq i \leq n$). Any span ${S'}$ = ${[e_p, e_{p+1}, ..., e_{p+k}]}\subseteq$ ${S}$, where $0 \leq p \leq p+k \leq n$, corresponds to a sub-sequence ${\Lambda'}$ of ${\Lambda}$ (i.e., ${v_p\rightarrow v_{p+1}\rightarrow ...\rightarrow v_{p+k}}$). Moreover, an eligible span ${S'}$ also corresponds to a connected substructure ${T'}$ of ${T}$, which meet the following 2 requirements:
\begin{itemize}
    \item there is at most one node ${v_i\in \Lambda'}$ which is the child node of node ${v\in \Lambda \backslash \Lambda'}$\footnote{If there is no such node, we specifiy that the first node in the in-order traversal sequence is ${v_i}$.};
    \item there is at most one node ${v_o\in \Lambda'}$ which is the parent node of node ${v\in \Lambda \backslash \Lambda'}$;
\end{itemize}
Note that each node in the tree $T$ has one parent node and at least one child node. Specially, the parent node of the root node and the child node(s) of the leaf node(s) are special imaginary nodes.
\end{myDef}

\section{Proofs for our theoretical insights}
In this Section, we present the proofs for our theoretical insights (Section \uppercase\expandafter{\romannumeral4}).
\begin{myLemma}
\label{lemma:strong_convexity}
\textbf{(Strong convexity of $h$)}
Given a vocabulary $V$ and a specific target index $y\in [0,V)$, let $p\in(0,1)$ represent prediction probability in the $y$-th dimension output by the sequence model and the negative likelihood $h(p) =-log(p) \in \mathbb{R}$ denote the optimization objective function of $p$. $h(p)$ is $\lambda(=4)$-strongly convex for on the feasible region of $p$. 
\end{myLemma}
\begin{myProof}
For any $p_1,p_2\in(0,1)$, we have the following derivations:
\small
\begin{align*}
    &\frac{(\nabla h(p_1)-\nabla h(p_2))(p_1-p_2)}{\lVert p_1-p_2 \rVert^2}=\frac{1}{p_1\cdot p_2}\geq \frac{4}{(p_1+p_2)^2}>4.
\end{align*}
\normalsize
That is $\nabla h(p_1)-\nabla h(p_2))(p_1-p_2)\ge\lambda(=4)\cdot\lVert p_1-p_2 \rVert^2$, which finishes the proof.
\end{myProof}

\begin{myAssume}
\label{lemma:orbit_average}
\textbf{(Unbiased on average)} Let $x$ represent a sample from the training set $D$, $f_\theta$ represent the neural network mapping function and $g$ represent a transform operation sampled from the group $\mathcal{G}$. We assume that: $\mathbb{E}_{g\in\mathcal{G}}[f_\theta(g\circ x)] = f_\theta(x)$.
\end{myAssume}

\begin{myTheo}
\label{lab:theo1}
Let $h$ denotes the negative likelihood loss function: $h(p) =-log(p)$.
We have the following inequality:
$
\mathbb{E}_{x_s,y_s}[h(p_\theta(y_*|y_s,x_s\oplus x_*)) + 2\sum_{g\in\mathcal{G}} \lVert p_\theta(y_*|g \circ y_s, (g\circ x_s)\oplus x_*)-p_\theta(y_*|y_s,x_s\oplus x_*)\rVert_2^2] \leq
\mathbb{E}_{x_s,y_s}\mathbb{E}_{g}[p_\theta(y_*|g \circ y_s,(g\circ x_s) \oplus x_*)]
$,
where the expression on the left hand side represents the original seq-to-seq loss function $\mathbb{E}_{x_s,y_s}[h(p_\theta(y_*|y_s,x_s\oplus x_*))$ with an additional regularization term $\sum_{g\in\mathcal{G}} \lVert p_\theta(y_*|g \circ y_s, (g\circ x_s)\oplus x_*)-p_\theta(y_*|y_s,x_s\oplus x_*)\rVert_2^2]$ and the expression on the right hand side represents the seq-to-seq loss function expected on the augmented dataset $\mathbb{E}_{x_s,y_s}\mathbb{E}_{g}[p_\theta(y_*|g \circ y_s,(g\circ x_s)\oplus x_*)]$.
\end{myTheo}
\begin{myProof}
Let $x_s$($y_s$) represents the original surrounding text in the input (output) and $g\in\mathcal{G}$ represents a substitution operation. Hence $(x_s\oplus x_*), (y_s\oplus y_*)$ represents an original example and $(g\circ x_s\oplus x_*), (g\circ y_s \oplus y_*)$ represents an augmented example.
Considering the loss objective function on $(g\circ x_s\oplus x_*), (g\circ y_s\oplus y_*)$, $h(p_\theta(y_*|g \circ y_s, (g\circ x_s)\oplus x_*))$, we have that (according to Lemma~\ref{lemma:strong_convexity}):
\small
\begin{align*}
    & h(p_\theta(y_*|g \circ y_s,(g\circ x_s)\oplus x_*)) \geq h(p_\theta(y_*|y_s,x_s\oplus x_*)) \\
    &+ \nabla h(p_\theta(y_*|y_s,x_s\oplus x_*)(p_\theta(g \circ y_s,(g\circ x_s)\oplus x_*)-p_\theta(y_*|y_s,x_s\oplus x_*)) \\
    &+ \frac{\lambda}{2}\lVert p_\theta(g \circ y_s,(g\circ x_s)\oplus x_*)-p_\theta(y_*|y_s,x_s\oplus x_*) \rVert_2^2.
\end{align*}
\normalsize
Through computing the expectation over the substitution operation $g$ and the surrounding text $(x_s,y_s)$ on both two sides, we induce that:
\small
\begin{align*}
    & \mathbb{E}_{g,x_s,y_s}[h(p_\theta(y_*|g \circ y_s,(g\circ x_s)\oplus x_*))] \geq \mathbb{E}_{x_s,y_s}[h(p_\theta(y_*|y_s,x_s\oplus x_*))] \\
    &+ \mathbb{E}_{g,x_s,y_s}[\nabla h( p_\theta(y_*|y_s,x_s\oplus x_*))(p_\theta(y_*|g \circ y_s,(g\circ x_s)\oplus x_*)\\
    &-p_\theta(y_*|y_s,x_s\oplus x_*))] \\
    &+ \mathbb{E}_{g,x_s,y_s}[\frac{\lambda}{2}\lVert p_\theta(y_*|g \circ y_s,(g\circ x_s)\oplus x_*)-p_\theta(y_*|y_s,x_s\oplus x_*) \rVert_2^2].
\end{align*}
\normalsize
We denote that $\nabla h( p_\theta(y_*|y_s,x_s\oplus x_*))(p_\theta(y_*|g \circ y_s,(g\circ x_s)\oplus x_*)
    -p_\theta(y_*|y_s,x_s\oplus x_*))\triangleq \Delta_1$ and $p_\theta(y_*|g \circ y_s,(g\circ x_s)\oplus x_*)] - p_\theta(y_*|y_s,x_s\oplus x_*)\triangleq \Delta_2$.
With Assumption~\ref{lemma:orbit_average}, we have the following derivations:
\small
\begin{align*}
    & \mathbb{E}_{g,x_s,y_s}[\Delta_1]=\mathbb{E}_{x_s,y_s}[\nabla h(x_s,y_s,x_*)(\mathbb{E}_{g}[\Delta_2])]= 0.
\end{align*}
\normalsize
Then we derive that:
\small
\begin{align*}
& \mathbb{E}_{g,x_s,y_s}[h(p_\theta(y_*|g \circ y_s,(g\circ x_s)\oplus x_*))] \\
& \geq \mathbb{E}_{x_s,y_s}[h(p_\theta(y_*|y_s,x_s\oplus x_*))] \\
& + \mathbb{E}_{g,x_s,y_s}[\frac{\lambda}{2}\lVert p_\theta(y_*|g \circ y_s,(g\circ x_s)\oplus x_*)-p_\theta(y_*|y_s,x_s \oplus x_*) \rVert_2^2].
\end{align*}
\normalsize
which finishes the proof.
\end{myProof}

\begin{myCorollary}
In the ideal conditions, where the training loss converges to zero and our compositional data augmentation fully substitutes the surrounding context $(x_s,y_s)$ for every $(x_*, y_*)$, we can disentangle the language modeling probability: $p_\theta(y_s\oplus y_* | x_s\oplus x_*) = p_\theta(y_s|x_s)\cdot p_\theta(y_*|x_*)$. 
\end{myCorollary}
\begin{myProof}
Following previous notations ($y_s=(y_s^1,y_s^2)$, where $y_s^1$ is ahead of $y_*$ and $y_s^2$ is behind of $y_*$.) and the formula of the conditional probability: we have $p_\theta(y_s\oplus y_* | x_s\oplus x_*) = p_\theta(y_{s}^1|x_s\oplus x_*)p_\theta(y_*|y_s^1,x_s\oplus x_*)p_\theta(y_s^2|y_*,y_s^1,x_s\oplus x_*)$. When the training loss converge to zero, we have that the regularization term introduced in the Theorem~\ref{lab:theo1} takes the value of zero: $\lVert p_\theta(y_*|g\circ y_s^1, (g\circ x_s^1)\oplus x_*) - p_\theta(y_*|y_s^1, x_s^1\oplus x_*)\rVert_2^2=0$. In other words, $p_\theta(y_*|g\circ y_s^1, (g\circ x_s^1)\oplus x_*) = p_\theta(y_*|y_s^1, x_s^1\oplus x_*)$ holds for any substitution operation $g$. If our compositional data augmentation can fully substitute the surrounding context $(x_s,y_s)$ of the component $(x_*,y_*)$, we can derive that the condition probability $p_\theta(y_*|y_s^1, x_s\oplus x_*)$ is independent to the selection $(x_s,y_s)$: $p_\theta(y_*|y_s^1, x_s\oplus x_*)=p_\theta(y_*|x_*)$. Similarly, we have $p_\theta(y_s^1|x_* \oplus x_s)=p_\theta(y_s^1|x_s),p_\theta(y_s^2|y_*,y_s^1,x_s\oplus x_*)=p_\theta(y_s^1,x_s)$ (Considering $(x_*,y_*)$ as the surrounding context of $(x_s,y_s)$). Hence we derive that:
\begin{align*}
& p(y_s \oplus y_* | x_s \oplus x_*) \\
& = p_\theta(y_{s}^1|x_s\oplus x_*)p_\theta(y_*|y_s^1,x_s\oplus x_*)p_\theta(y_s^2|y_*,y_s^1,x_s\oplus x_*) \\
& = p_\theta(y_s^1|x_s)\cdot p_\theta(y_*|x_*) \cdot p_\theta(y_s^2|y_s^1,x_s)\\
&=p_\theta(y_s^1,y_s^2|x_s)\cdot p_\theta(y_*|x_*)=p_\theta(y_s|x_s)\cdot p_\theta(y_*|x_*), 
\end{align*}
which ends the proof.
\end{myProof}

\begin{myTheo}
Given a substitution operation group $\mathcal{G}$, define $\hat{l}$ as $\hat{l}(x_i,y_i)=\max_{g\in\mathcal{G}}(l(h(g\circ x), g\circ y))$. We have that the expected risk for the compositional generalization $R_{comp}(h)$ is upper bounded by $R_\mathcal{S}(\hat{l}\circ h) + 2\mathcal{R}_\mathcal{S}(\hat{l}\circ h) + 3\sqrt{\frac{\log\frac{2}{\delta}}{2n}} + L\mathbb{W}_1(\mathcal{X}_{aug},\mathcal{X}_{comp})$.
\end{myTheo}
\begin{myProof}
    We have the following derivations:
    \small
    \begin{align*}
    & R_{comp}(h)=R_{aug}(h) + (R_{comp}(h)-R_{aug}(h)) \\
    & \leq R_{aug}(h) + L\mathbb{W}_1(\mathcal{X}_{aug},\mathcal{X}_{comp}) \\
    & = \mathbb{E}_{(x,y)\sim\mathcal{X}_{aug}}[l(h(x),y)] + L\mathbb{W}_1(\mathcal{X}_{aug},\mathcal{X}_{comp}) \\
    & \leq \mathbb{E}_{(x,y)\sim\mathcal{X}_{aug}}[\max_{g\in\mathcal{G}}l(h(g\circ x),g\circ y)] + L\mathbb{W}_1(\mathcal{X}_{aug},\mathcal{X}_{comp}) \\
    & = R_{aug}(\hat{l}\circ h) + L\mathbb{W}_1(\mathcal{X}_{aug},\mathcal{X}_{comp}) \\
    & \leq \underbrace{R_{\mathcal{S}}(\hat{l}\circ h)}_{\text{empirical risk}} + \underbrace{2\mathcal{R}_\mathcal{S}(\hat{l}\circ h)}_{\text{rademacher complexity}} + 3\sqrt{\frac{\log\frac{2}{\delta}}{2n}}+ \underbrace{L\mathbb{W}_1(\mathcal{X}_{aug},\mathcal{X}_{comp})}_{\text{distribution distance}}. 
    \end{align*}
    The last step is derived from Mohri et al.~\cite{10.5555/2371238}. This finishes the proof.
\end{myProof}

\section{Detailed Experimental Settings}
\label{appendix:experimental settings}
In this section, we detailedly describe the training details of our models in our framework (up-stream LCS Augmentor and down-stream neural seq-to-seq model), the selection of hyper-parameters in our Algorithms(CompSub and LCS), the generation configurations for the LCS-ICL inference of LLMs (LLaMA2-13B~\cite{touvron2023llama2openfoundation} and LLaMA3-8B~\cite{dubey2024llama3herdmodels}) and other details. 
\subsection{LCS Augmentor}
For both of SCAN and COGS experiments, we use an two layer bidirectional LSTM (with 128 hidden units and an embedding size of 128, a dropout rate of 0.5) as our sequence encoder. We separately use an embedding layer with an embedding size of 512 for the embedding module for spans to be substituted out and another embedding layer with an embedding size of 512 for the embedding module for spans to be substituted in. We use (cosine-similarity$\cdot$2) $ \in [-2,2]$ as all of our similarity functions in LCS augmentor. We set all of the temperatures for gumbel-softmax sampling in LCS augmentor to 1.
Besides, we use a Adam optimizer~\cite{Kingma2014AdamAM} to optimize our LCS augmentor with an learning rate of 1e-3.
The above hyper-parameters are commonly used for LSTM-based models in NLP community and hence we did not spend extra efforts to tune them in our experiments.
\label{append:model_augmentor}
\subsection{Neural Language Models}
\label{append:model_parser}
We keep this part of hyper-parameters aligned with previous baselines.
For \emph{jump} and \emph{around right} splits of SCAN and COGS experiments, we keep the hyperparameters of our LSTM in align with \cite{lexlearn, lexsym, mutual}. We use a 2-layer encoder-decoder LSTM (with attention~\cite{attention} and copy~\cite{copy} mechanisms) with 512 hidden units and an embedding size of 512, a droupout rate of 0.4. For \emph{MCD}1, \emph{MCD}2 and \emph{MCD}3 splits of SCAN experiments, the hyperparameters of our LSTM are adopted form \cite{geca}. We use a 1-layer bidirectional encoder-decoder LSTM (with attention and copy mechanisms) with 512 hidden units and an embedding size of 64, a droupout rate of 0.5. For all of these above experiments, we train our model with an Adam optimizer with an initial learning rate of 1e-3. We use an ReduceLROnPlateau scheduler (implemented in PyTorch) with a scale factor of 0.5 to automatically reduce our learning rate. We set all of the batch size to 128.

For GeoQuery tasks, in align with SUBS~\cite{subs}, we also directly use OpenNMT~\cite{opennmt} to implement our LSTM-based model with attention and copy mechanisms and we utilize fairseq~\cite{fairseq} to implement our BART-based model. For LSTM-based experiments, we use one-layer bidirectional LSTM in the encoder side and one-layer unidirectional LSTM in the decoder side. We use dropout with a rate of 0.5 and Adam optimizer with a learning rate of 1e-3. We use MLP attention and directly use the attention scores as copying scores and we set the batch size for experiments based on LSTM to 64. For BART-based experiments, we use BART-base models updated by Adam optimizer with a learning rate of 1e-5. We set the rate for both dropout and attention dropout to 0.1 and we use label smoothing with a rate of 0.1. We set the batch size for all of the experiments based on BART to 1024 tokens. Besides, we set the rate of the weight-decay to 0.01. 
\subsection{Hyper-parameters in CompSub (Algorithm 1)}
For \emph{jump} and \emph{around right} splits of SCAN and GeoQuery experiments, we set the iterative depth $K$ in CompSub augmentation scheme to 1. For \emph{MCD} splits of SCAN experiments, we set the iterative depth $K$ in CompSub augmentation scheme to 2. For COGS experiments, we set the iterative depth $K$ in CompSub augmentation scheme to 4. For SCAN experiments, we set the number of generated examples $N$ (without de-duplicating) to 1e5. For COGS experiments, we set the number of generated examples $N$ (without de-duplicating) to 4e5. For GeoQuery experiments, we simply searching for every potential augmentations in the training set (because the training set for GeoQuery contains merely 519 examples, we try to make the best use of each example.), and the size of augmented set is shown in Table \uppercase\expandafter{6}. Following ~\cite{jia_liang, qiu_csl}, we also ensure approximately equal number of the original examples and the augmented examples being used for training in CompSub experiments, giving consideration to both of i.i.d. generalization and compositional generalization.

We decide the iterative depth $K$ through observing that from which iteration there are nearly no more novel data generated. For $N$, we simply set a number which is large enough compared with the size of the original dataset, and then we de-duplicate the augmented dataset. 
\subsection{Hyper-parameters in LCS (Algorithm 2)}
One crucial hyper-parameter in Training LCS framework is the warm-up epochs / update steps. In most cases, we need to set an appropriate value to warm-up update steps to guarantee the down-stream sequence model to be fully aware of the distribution (hardness) of the original training examples while not over-fit to them.
For most of our experiments(\emph{jump}, \emph{around right}, \emph{MCD1} and \emph{MCD2} splits of SCAN experiments, COGS experiments), we set the warm-up epoch to 5, and then we alternatively train the up-stream module and down-stream module in the LCS framework to 150 epochs in total. For \emph{MCD2} split of SCAN experiments, we first train our neural seq-to-seq model for 80 epochs, and then we alternatively train the up-stream LCS augmentor and the down-stream neural seq-to-seq model for 70 epochs\footnote{In our initial experiments, we found that LCS method only slightly works on the \emph{MCD2} split of SCAN dataset when using 1 layer LSTM-based model as the down-stream sequence model. However, in the following experiments in Table \uppercase\expandafter{\romannumeral9}, we found that it works well on other 2 down-stream sequence models (we set warm-up epoch number to 5 for other down-stream seq-to-seq models).}.
For experiments with LCS framework, we set the number of sampled actions $T$ for each example to 4. All of this part of hyper-parameters are decided by cross-validation.
\subsection{Generation Configurations of LLMs}
We conduct all of the LCS-ICL experiments with both LLaMA2-13B~\cite{touvron2023llama2openfoundation} and LLaMA3-8B~\cite{dubey2024llama3herdmodels}. For LLaMA2-13B, we use the version of meta-llama/Llama-2-13b-hf\footnote{\url{https://huggingface.co/meta-llama/Llama-2-13b-hf}}; For LLaMA3-8B, we use the version of meta-llama/Meta-Llama-3-8B-Instruct\footnote{\url{https://huggingface.co/meta-llama/Meta-Llama-3-8B-Instruct}}. We use the default generation configurations for both two LLMs: (1) for LLaMA2-13B, the generation configuration is:
\begin{lstlisting}
GEN_CONFIGS["llama2-13b"]={
  "bos_token_id": 1,
  "do_sample": True,
  "eos_token_id": 2,
  "pad_token_id": 0,
  "temperature": 0.6,
  "max_length": 100,
  "top_p": 0.9,
  "transformers_version": "4.31.0.dev0"
}
\end{lstlisting}
(2) for LLaMA3-8B, the generation configuration is:
\begin{lstlisting}
GEN_CONFIGS["llama3-8b"]={
  "bos_token_id": 128000,
  "do_sample": True,
  "eos_token_id": [128000,128009],
  "temperature": 0.6,
  "max_length": 300,
  "top_p": 0.9,
  "transformers_version": "4.31.0.dev0"
}
\end{lstlisting}
\subsection{Other Training and Inference Details}
We conduct all of our training experiments on eight NVIDIA GeForce RTX2080Ti GPUs. For \emph{jump} and \emph{around right} splits of SCAN, COGS and GeoQuery, we select our model for testing with the best development accuracy.
We conduct all of our LLM In-Context Learning inference experiments on two NVIDIA A100 GPUs. Note that we use 16 bit quantization for both LLaMA2-13B and LLaMA3-8B.
For all \emph{MCD} splits of SCAN, we use the train/dev/test splits from the original paper~\cite{MCD}\footnote{The official github repo is \url{https://github.com/google-research/google-research/tree/master/cfq\#scan-mcd-splits}, and one can download the dataset from \url{https://storage.cloud.google.com/cfq_dataset/scan-splits.tar.gz}}, we also select our model for testing with the best accuracy on dev set.

\section{Others}
\subsection{Pseudo codes for the LCS-ICL algorithm}
We present the pseudo codes for the LCS-ICL algorithm in Algorithm~\ref{algo:lcs_icl}: we conduct the coarse screening in line $6\sim22$, the demonstration augmentation in line $24$ and the fine screening in line $28\sim37$.

\begin{algorithm}[t]
	\caption{\textbf{LCS-ICL}}
        \label{algo:lcs_icl}
	\KwIn{Original demonstration pool ${\mathcal{D}}$,\\Query input $x_{q}$, \\The number of the in-context demonstrations $k$, \\Large language model $f_\theta(\cdot)$.
    }
	\KwOut{$k$ demonstrations $\mathcal{C}=\{(inp_i,out_i)\}_{i=1}^k$.}  

        $target=\text{Tokenize}(x_{q})$;\Comment{\textcolor{blue}{$\text{Tokenize}(x)$ will return a set of tokens that are covered in $x$.}}
        
        $uncover=\text{Tokenize}(x_{q})$;

        $\mathcal{D}'=\mathcal{D}$;

        $\mathcal{S}=\{\}$;

        $step = 0$;
        
        \While{${step} < \lceil k/2 \rceil$ or $uncover \neq \emptyset$}
        { 
            $\text{CoverScore}=dict()$;  \Comment{\textcolor{blue}{Coarse Screening}}
            
            \If{$step>k$}{ 
                break;
            }
            
            \For{$(x_i,y_i)\in\mathcal{D}'$}
            {
                $toks=\text{Tokenize}(x_i)$;

                $\text{CoverScore}[x_i]=0$;
                
                \For{tok $\in toks$}
                {
                    \If{tok $\in uncover$}
                    {
                        $\text{CoverScore}[x_i] = \text{CoverScore}[x_i]+2$;

                        $uncover=uncover-\{tok\}$
                    }
                    \ElseIf{tok $\in target$}
                    {
                        $\text{CoverScore}[x_i] = \text{CoverScore}[x_i]+1$;
                    }
                }
            }
            $s=\mathop{\text{argmax}}\limits_i(\text{CoverScore}[x_i])$;

            $\mathcal{C}=\mathcal{C}+\{(x_s,x_s)\}$;
            
            $\mathcal{D}'=\mathcal{D}'-{(x_s,x_s)}$;

            $step = step +1$;
        } 

        $m = step$;
        
        $\mathcal{D}^*=\text{CompSub}(\mathcal{C})$ \Comment{\textcolor{blue}{Demonstration Augmentation}}

        $\text{Prefix}=\text{“"}$;
            
        \For{$(x_i,y_i)\in\mathcal{S}$}
        {
            $\text{Prefix}+=\text{“input:"}+x_i+\text{“;output:"}+y_i$;
        }

        \For{${step}\leftarrow$ $m+1$ to $k$}
        {
            $\text{LCS-Score}=dict()$;\Comment{\textcolor{blue}{Fine Screening}}

            $\mathcal{S} = \text{random.sample}(\mathcal{D}^*,100)$;
            
            \For{$(x_{cand},y_{cand})\in \mathcal{D}^*$}
            {
                $\text{Prompt}=\text{Prefix}+\text{“input:"}+x_{cand}+\text{“;output:"}+y_{cand}$;

                $\text{LCS-Score}[x_{cand}]=\text{PPL}(f_\theta(\text{Prompt}), y_{cand})$;
                
            }
            $s=\mathop{\text{argmax}}\limits_{i}(\text{LCS-Score}[inp_{i}])$;

            $C=C+\{(inp_s,out_s)\}$;
            
            $\mathcal{D}^*=\mathcal{D}^*-\{(inp_s,out_s)\}$;

            $\text{Prefix}=\text{Prefix}++\text{“input:"}+inp_s+\text{“;output:"}+out_s$;
        }
        
	\Return{$C$;}
\end{algorithm}
\subsection{Calculating the Wasserstein distance between the augmented data distribution and the compositional testing data distribution}
In our paper, we argue that the value of the expression $L\mathbb{W}_1(\mathcal{X}_{aug},\mathcal{X}_{comp})$ depends on the inherent ability of the data augmentation scheme: in comparison with previous compositional data augmentation scheme (e.g., SUBS~\cite{subs} and LexSym~\cite{lexsym}), CompSub can achieve a relatively low 1-Wasserstein distance between the augmented training data distribution with the true compositional testing data distribution.
We calculate the 1-Wasserstein distances between the augmented training data (generated by our algorithm CompSub, LexSym~\cite{lexsym} and SUBS~\cite{subs}) and the true compositional testing data. We conduct experiments on the SCAN dataset (MCD1 split) and the COGS dataset. The results are shown in Table~\ref{tab:wasserstein}. We observe that CompSub indeed achieves lower 1-Wasserstein distance (to the real compositional testing data) compared with previous baseline augmentation methods (LexSym and SUBS).
Note that for the experiments on the SCAN dataset, since the vocabulary set is small, we use the 3-gram language modeling method to encode each sentence; for the experiments on the COGS dataset, since the vocabulary set is large, we use a pre-trained sentence embedding model (paraphrase-MiniLM-L6-v2\footnote{\url{https://huggingface.co/sentence-transformers/paraphrase-MiniLM-L6-v2}}) to encode each sentence. To calculate the wasserstein distance, we use the off-the-shelf POT (Python Optimal Transport) package~\cite{flamary2021pot}.

\begin{table}[t]
\caption{
1-Wasserstein distance between the augmented training data distribution and the real compositional testing data distribution.
}
\centering
\resizebox{0.48\textwidth}{!}{
\begin{tabular}{l|ccc}
\hline
\toprule
\textbf{Dataset} & SUBS~\cite{subs} & LexSym~\cite{lexsym} & CompSub (\textit{Ours})  \\
\midrule
SCAN~\cite{scan} & -- & 2.37  & 2.10 \\
COGS~\cite{cogs} & 6.71  & 6.67 & 6.49 \\
\bottomrule
\end{tabular}
}
\label{tab:wasserstein}
\end{table}

\end{document}